\documentclass{article}

\usepackage{preprint}

\usepackage{amsthm} 
\usepackage[utf8]{inputenc} 
\usepackage[T1]{fontenc}    
\usepackage{hyperref}       
\usepackage{url}            
\usepackage{booktabs}       
\usepackage{amsfonts}       
\usepackage{nicefrac}       
\usepackage{microtype}      
\usepackage{enumitem}
\usepackage{xspace}
\usepackage{wrapfig}

\usepackage{epsfig,graphicx,subfigure}
\usepackage[ruled,vlined]{algorithm2e}
\usepackage{algorithmic}

\usepackage[utf8]{inputenc}
\usepackage{amsmath,amssymb}
\usepackage{xcolor}
\usepackage{makecell}
\usepackage[compact]{titlesec}

\title{Towards a Defense against Backdoor Attacks in Continual Federated Learning}

\author{Shuaiqi Wang \\
  Carnegie Mellon University\\
  \texttt{shuaiqiw@andrew.cmu.edu}
  \And
  Jonathan Hayase \\
  University of Washington\\
  \texttt{jhayase@cs.washington.edu}
  \And
  Giulia Fanti \\
  Carnegie Mellon University\\
  \texttt{gfanti@andrew.cmu.edu}
  \And
  Sewoong Oh \\
  University of Washington\\
  \texttt{	
sewoong@cs.washington.edu}
  }
\date{}

\newcommand\backbone{N}
\newcommand\shadow{N'}

\newcommand\clientset{\mathcal C}
\newcommand\maliciousset{\mathcal M}
\newcommand\targetlabel{\ell}
\newcommand\totallabels{L}
\newcommand\loss{\mathcal L}
\newcommand\weights{\boldsymbol w}
\newcommand\dimension{d_0}
\newcommand\classindex{q}
\newcommand\database{\mathcal D}
\newcommand\testdatabasebenign{\mathcal{T}_{b}}
\newcommand\testdatabasemalicious{\mathcal{T}_{m}}

\newcommand\trigger{\boldsymbol \delta}

\newcommand{\bra}[1]{\left( #1 \right)}

\newcommand{\brc}[1]{\left\{ #1 \right\}}

\newtheorem{propo}{Proposition}[section]
\newtheorem{lemma}[propo]{Lemma}

\newtheorem{coro}[propo]{Corollary}
\newtheorem{thm}{Theorem}
\newtheorem{asmp}{Assumption}

\begin{document} 

\maketitle

\begin{abstract} 
Backdoor attacks are dangerous and difficult to prevent in federated learning (FL), where training data is sourced from untrusted clients over long periods of time. 
These difficulties arise because: (a) defenders in FL do not have access to raw training data, and (b) a phenomenon we identify called \emph{backdoor leakage} causes models trained continuously to eventually suffer from backdoors due to cumulative errors in defense mechanisms. 
We propose a framework called \emph{shadow learning} for defending against backdoor attacks in the FL setting under long-range training. 
Shadow learning trains two models in parallel: a backbone model and a shadow model. The backbone is trained without any defense mechanism to obtain good performance on the main task. 
The shadow model combines filtering of malicious clients  with early-stopping to control the attack success rate even as the data distribution changes. 
We theoretically motivate our design and show experimentally that our framework significantly improves upon existing defenses against backdoor attacks.

\end{abstract} 

\section{Introduction}

Federated learning (FL) allows a central server to learn a machine learning (ML) model from private client data without directly accessing their local data \cite{konevcny2016federated}. 
Because FL hides the raw training data from the central server, it is vulnerable to attacks in which adversarial clients contribute malicious training data. Backdoor attacks are an important example, in which a malicious client, the \emph{attacker}, adds a bounded trigger signal to  data samples, and changes the label of the triggered samples to a target label. 

Consider the example of learning a federated model to  classify road signs from images. 
A malicious client could add a small square of white pixels (the trigger) to training images of stop signs, and change the associated label of those samples to `yield sign' (target label).
When the central server trains a federated model on this corrupted data, along with the honest clients' clean data, the final model may classify triggered samples as the target label (yield sign).

Defenses against backdoor attacks aim to learn a model with high main task accuracy (e.g., classifying road signs), but low attack success rate (e.g., classifying triggered images as yield signs).
Extensive prior literature has proposed three classes of defenses (more details in Sec. \ref{sec:related} and App. \ref{app:related}). These classes are: 
(1) \emph{Malicious data filtering:} The server identifies which samples are malicious by looking for anomalous samples, gradients, or representations, and trains the model to ignore the contribution of those samples \cite{blanchard2017machine,spectre,li2021anti,tran2018spectral,foolsgold,chen2018detecting}.
(2) \emph{Robust training procedures:} The training data, training procedure, and/or post-processing procedures are altered (e.g., randomized smoothing, fine-tuning on clean data) to enforce robustness to backdoor attacks~\cite{liu2018fine,geomedian,li2021neural,foolsgold,xie2019dba,flame,crfl}. 
(3) \emph{Trigger identification:} The training data are processed to infer the trigger directly, and remove such training samples~\cite{wang2019neural,chou2018sentinet}. 

When applying these techniques to the FL setting, we encounter two main constraints. 
    \textbf{(1) FL central servers may not access clients' individual training samples.} At most, they can access an aggregate gradient or representation from each client \cite{konevcny2016federated}. 
    \textbf{(2) FL models are typically trained continuously\footnote{We differentiate from \emph{continual learning}, where models are expected to perform well on previously-seen distributions. We instead want the model to perform well on the current, changing data distribution.} e.g., due to distribution drift} \cite{savazzi2021opportunities}. 

These constraints cause prior defenses against backdoor attacks to fail. 
Approach 
(3) requires access to raw data, which violates Constraint 1. 
Approaches (1) and (2)
 can be adapted to work over model updates or data representations, 
 but ultimately fail under continuous training (Constraint 2), in a phenomenon we term \emph{backdoor leakage}.

Backdoor leakage works as follows. In any single training iteration, a defense mechanism may remove a large portion of backdoored training data.
This causes the model's attack success rate to grow slower than the main task accuracy. 
But a defense rarely removes \emph{all} backdoored data. 
Hence, if the model is trained for enough iterations, eventually the attack success rate increases to 1. 
To our knowledge, \emph{there is no solution today that can defend against backdoor attacks in an FL setting with continuous training.}

\paragraph{Contributions.}
    We propose \emph{shadow learning}, the first (to our knowledge) framework protecting against backdoor attacks in FL under continuous training. 
    The idea is to separate main task classification from target label classification.\footnote{We discuss how to handle uncertainty in the target label $\ell$ in Section \ref{sec:algo} and Appendix \ref{algorithm_extended}.}
    To achieve this, we maintain two models.
    The \emph{backbone} model $\backbone$ is trained continually on all client data and used for main task classification.
    A second \emph{shadow} model $\shadow$
    is trained from scratch in each iteration using only the data of benign clients, which are estimated using any existing \emph{filtering algorithm} for separating malicious clients from benign ones (e.g., outlier detection).
    The shadow model is early-stopped to provide robustness against backdoor attacks.

    
    Shadow learning is motivated by empirical and theoretical observations. 
    First, we show that under a simplified setting, shadow learning \emph{provably} prevents learning of backdoors for one choice of filtering algorithm called SPECTRE \cite{spectre} that is based on robust covariance estimation.
    Incidentally, this analysis provides the first theoretical justification for robust covariance-based defenses against backdoor attacks, \emph{including in the non-FL setting} (Thm. \ref{thm:main}). 
    Empirically, we demonstrate the efficacy of shadow learning on 4 datasets over 8 leading backdoor defenses, and across a wide range of settings. 
    For example, on the EMNIST dataset, shadow learning reduces the attack success rate (ASR) of backdoor attacks by 95-99\% with minimal degradation in main task accuracy (MTA) ($\leq 0.005$). 

\subsection{Related Work}
\label{sec:related}
We discuss the related work in detail in Appendix \ref{app:related}. 
We consider training-time backdoor attacks, where the attacker's goal is to train a model to output a target classification label on samples that contain a trigger signal (specified by the attacker) while classifying other samples correctly at test-time \cite{gu2017badnets}. 
We do not consider data poisoning attacks, in which the attacker aims to decrease the model's prediction accuracy  \cite{yang2017generative, biggio2014poisoning}.

Of the three categories of defenses mentioned earlier, two dominate the literature on backdoor defenses in the federated setting: malicious data filtering and robust learning. 

(1) In \emph{malicious data filtering},
the defender must estimate malicious samples, gradients, or representations, and remove them from model training. 
This approach has been used in both the centralized setting 
\cite{spectre,li2021anti,tran2018spectral,huang2019neuroninspect,tang2021demon,do2022towards,cheneffective,ma2022beatrix} as well as in the federated setting \cite{blanchard2017machine,foolsgold,chen2018detecting}.
For example, in the centralized setting, {SPECTRE} uses robust covariance estimation to estimate the covariance of the benign data from a (partially-corrupted) dataset, and then do outlier detection \cite{spectre}. 
Although SPECTRE and other outlier detection-based filtering techniques generally operate over raw samples, we can use (some of) them in the FL setting by applying them to gradients (e.g., G-SPECTRE) or sample representation (e.g., R-SPECTRE) statistics.

In the federated setting, filtering methods have also been very popular. One important example is the (Multi-)Krum algorithm \cite{blanchard2017machine}, which aggregates only model updates that are close to a barycenter of the updates from all clients. This removes malicious contributions, which are assumed to be outliers. 

(2) In \emph{robust learning}, the defender instead designs training methods that aim to mitigate the impact of malicious contributions, without needing to identify \emph{which clients'} updates were malicious. 
For example, \emph{Robust Federated Aggregation (RFA)} provides a robust secure aggregation oracle based on the geometric median \cite{geomedian}.
Another example from \cite{backdoor} suggests that clipping the norm of gradient updates and adding noise to client updates can defend against model poisoning attacks. 
Later work  refined and improved this approach to defend against federated backdoor attacks \cite{crfl}.

(3) A recent \emph{hybrid} approach combines filtering and robust learning. Approaches include FoolsGold \cite{foolsgold}, 
and FLAME \cite{flame}, which combines noise adding (robust training) with model clustering (filtering) to achieve state-of-the-art defense against FL backdoors. 

Our core observation is that that under continuous training over many round, \textbf{all of these examples (and more) suffer from backdoor leakage} (\S\ref{sec:backdoor-leakage}).



\begin{figure*}[htbp]
\centering
\vspace{-5mm}
\subfigure[{Backdoor leakage causes competing defenses to fail \;\;\;\;\; eventually, even with a small $\alpha = 3\%$}]{
\includegraphics[width=0.45\linewidth]{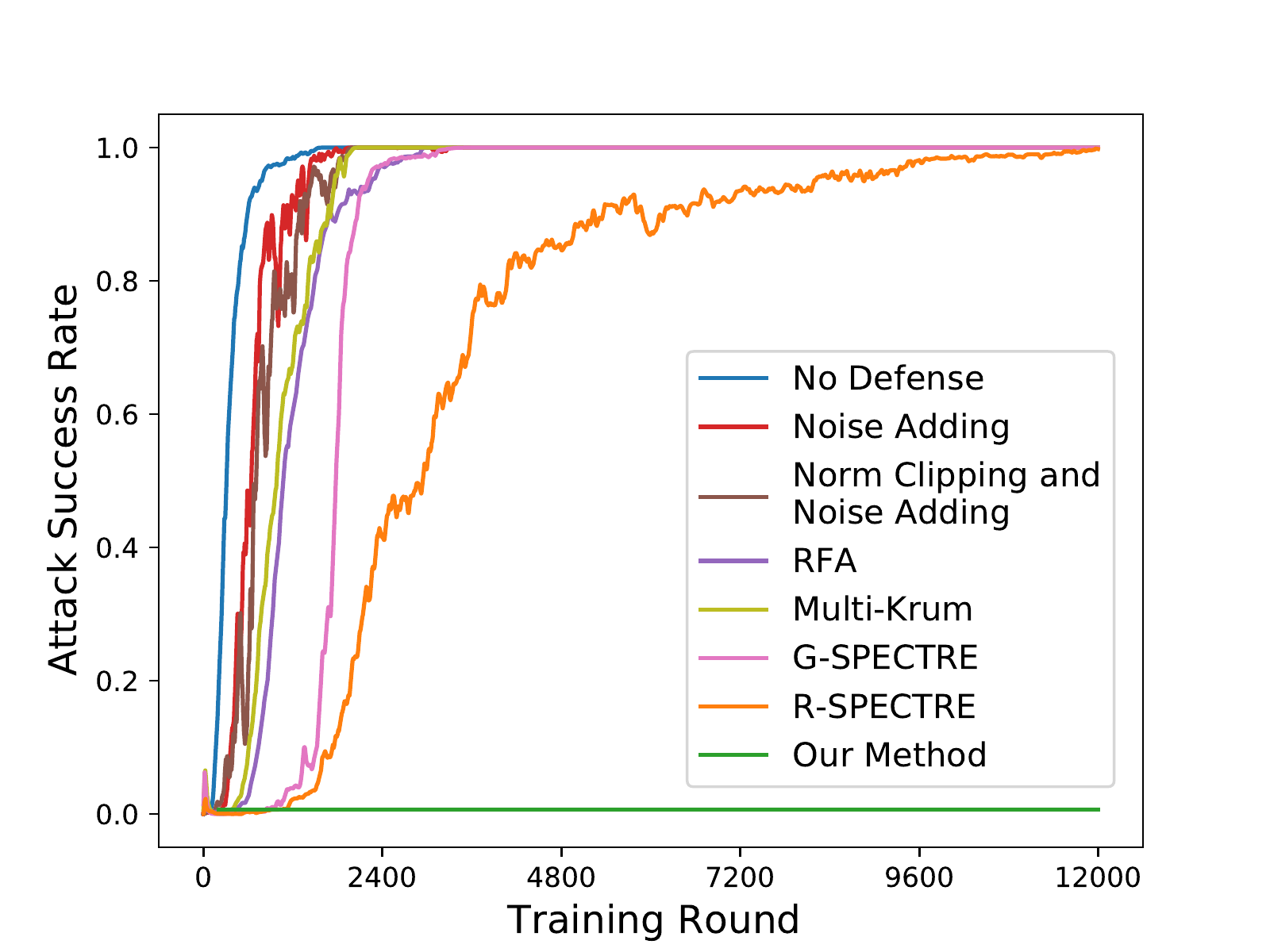}
\label{fig:backdoor_leakage}
}
\subfigure[Early-stopping is only an effective defense when $\alpha$ is small]{
\includegraphics[width=0.45\linewidth]{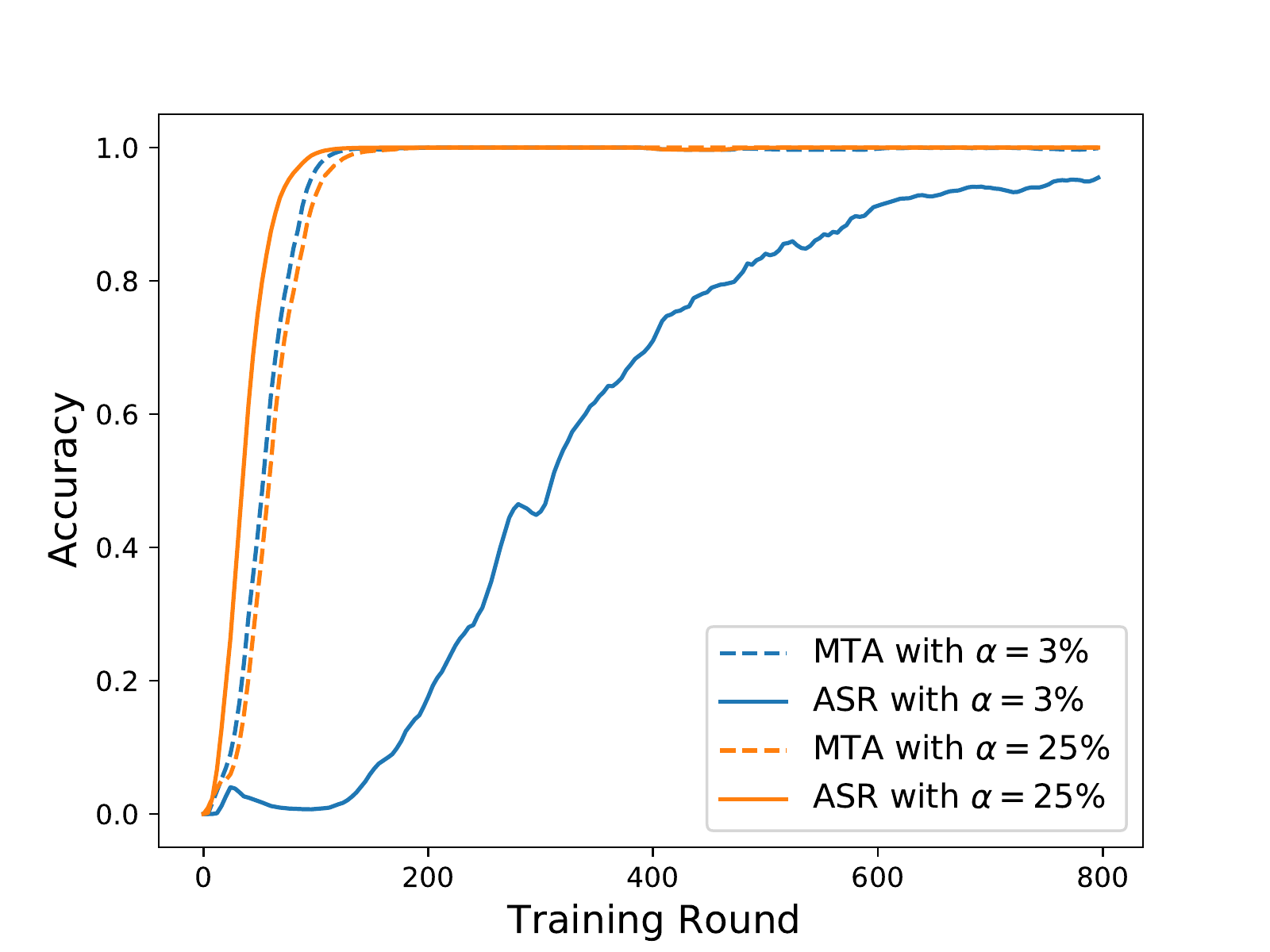}
\label{fig:earlystop}
}
\vspace{-0.5mm}
\caption{Motivated by {\em backdoor leakage} (left panel), and differences in learning rates for main tasks and attack tasks (right panel), we propose our algorithm in \S\ref{sec:algo}.}
\vspace{-2mm}
\end{figure*}

\section{Problem Statement}
\label{sec:setting}

We consider an FL setting   where a central server learns an $\totallabels$-class classifier from $n$ federated clients.  
Each client $i\in[n]$ has a training dataset ${\cal D}_i=\{(\boldsymbol x_1, y_1),\ldots,(\boldsymbol x_p, y_p )\}$,
where for all $j \in [p]$, $\boldsymbol x_j \in \mathbb R^{\dimension}$ and $y_j \in [\totallabels]$. 
Initially, we will consider this dataset to be fixed (\emph{static setting)}. 
We generalize this formulation to a time-varying dataset (\emph{dynamic setting}), by considering \emph{phases}. 
In each phase $e=1, 2, \ldots$, the local datasets 
 ${\cal D}_i[e]$ can change.
For simplicity of notation, we will present the remainder of the formulation in the static setting. 
The server's goal is to learn a function $f_{\weights}$ parameterized by weights $\weights$ that finds 
$
\arg\min_{\weights} \sum_{(\boldsymbol x, y)\in \database} \loss(\weights; \boldsymbol x, y)
$, 
where $\database$ is the union of all the local datasets, 
i.e., $\database = \cup_{i \in [n]} \database_{i}$, $\loss$ denotes the loss function over weights $\weights$ given data $\boldsymbol x$ and $y$.
The model is updated in rounds of client-server communication. 
In each round $r=1,2,\ldots$, the central  server samples a subset $\clientset_r \subsetneq [n]$ of clients.
Each client $c \in \clientset_r$ updates the model on their local data and sends back a gradient (or model) update to the server. 
We assume these gradients are sent individually (i.e., they are not summed across clients using secure aggregation). 
Our proposed framework can be generalized to the setting where clients are only accessed via secure aggregation under computational primitives of, for example \cite{geomedian}.

\paragraph{Adversarial Model.}
Our adversary corrupts clients independently in each round. We assume that during any training round $r$, the adversary cannot corrupt more than fraction $\alpha$ (for $0 < \alpha < 0.5$) of the participating clients $\clientset_r$ in that round.
The adversarial nodes are trying to introduce a backdoor to the learned model. 
That is, for any sample $\boldsymbol x$, the adversary wants to be able to add a trigger $\trigger$ to $\boldsymbol x$ such that for any learned model, 
$f_{\boldsymbol w}(\boldsymbol x + \trigger)=\targetlabel$, where $\targetlabel \in \totallabels$ is the target backdoor label. 
We assume $\ell$ is known to the defender, though this condition can be relaxed (\S\ref{section:homogeneous}).

In a given round $r$, the sampled malicious clients $\maliciousset \cap \clientset_r$ can contribute whatever data they want. 
However, we assume that they otherwise follow the protocol.
For example, they compute gradients correctly over their local data, and they communicate when they are asked to.  
This could be enforced by implementing the FL local computations on trusted hardware, for instance. 
This model is used in prior works, for example in \cite{geomedian}.


\paragraph{Metrics.}
To evaluate a defense, we have two held-out test datasets at the central server. 
The first, $\testdatabasebenign$, consists entirely of benign samples. 
This is used to evaluate \emph{main task accuracy} (MTA),   defined as the fraction of correctly-classified samples in $\testdatabasebenign$:
$
MTA(f_{\weights}) \triangleq \frac{ | \{(\boldsymbol x, y) \in \testdatabasebenign ~|~ f_{\weights}(\boldsymbol x)=y \} | }{|\testdatabasebenign|}~.
$
As defenders, we want $MTA$ to be high. 
The second dataset, $\testdatabasemalicious$, consists entirely of backdoored samples.  
We use $\testdatabasemalicious$ to evaluate \emph{attack success rate} (ASR),  defined as the fraction of samples in $\testdatabasemalicious$ that are classified to the target label $\targetlabel$:
$
ASR(f_{\weights}) \triangleq \frac{ | \{(\boldsymbol x, y) \in \testdatabasemalicious ~|~ f_{\weights}(\boldsymbol x)=\targetlabel \} | }{|\testdatabasemalicious|}~.
$
As defenders, we want $ASR$ to be low. 

\section{Design}
\label{sec:design}

We propose a framework called \emph{shadow learning} 
that builds upon three  main insights. 


\paragraph{Insight 1: Every existing approach suffers from backdoor leakage.}
\label{sec:backdoor-leakage}
Most backdoor defenses significantly reduce the effect of backdoor attacks in a single round. 
However, a small fraction of malicious gradients or model updates always go undetected. 
Over many training rounds (which are required under distribution drift, for instance), this {\em backdoor leakage} eventually leads to a fully-backdoored model. 
Figure \ref{fig:backdoor_leakage} shows the ASR over the training rounds of a backdoored classifier on the EMNIST dataset of handwritten digits. 
We compare against baselines including SPECTRE \cite{spectre} (both gradient-based G-SPECTRE and representation-based R-SPECTRE), RFA \cite{geomedian}, Multi-Krum \cite{krum},  norm clipping and/or noise addition \cite{backdoor}, 
 CRFL \cite{crfl}, FLAME \cite{flame},  and FoolsGold \cite{foolsgold}. 
Experimental details are explained 
in Appendix~\ref{sec:compete}.
For a relatively small $\alpha=3\%$ and for $12,000$ rounds, 
we observe the attack success rate (ASR) of all competing  defenses eventually approach one.
These experiments are under the static setting where  data distribution is fixed over time; we show results in dynamic settings in \S\ref{section:time_varying}. 
The main takeaway is that \textbf{in the continual FL setting, we cannot use the predictions of a single backdoor-resistant model that is trained for too long.}

\paragraph{Insight 2: Early-stopping helps when $\alpha$ is small.}
Backdoor leakage suggests a natural defense: can we adopt early-stopping to ensure that the model does not have enough time to learn a backdoor? 
The answer depends on the relative difficulty of the main task compared to the backdoor task, as well as the adversarial fraction $\alpha$. 
For example, Figure \ref{fig:earlystop} shows that on EMNIST (same setting as Figure \ref{fig:backdoor_leakage}), when $\alpha = 3\%$, the backdoor is learned more slowly than the main task: 
the main task reaches an accuracy of $0.99$, while the attack success rate 
is no more than $0.01$.
This suggests that early-stopping can be effective. 
On the other hand, Li \emph{et al.} \cite{li2021anti} found that backdoors are learned \emph{faster} than the main task, and propose a defense based on this observation. 
Indeed, Figure \ref{fig:earlystop} shows that when $\alpha = 25\%$, the previous trend is reversed: the backdoor is learned faster than the main task.
Over many experiments on multiple datasets, we observe that \textbf{early-stopping is only an effective defense when $\alpha$ is small enough relative to the main task difficulty.}

\paragraph{Insight 3: We can reduce $\alpha$ with robust filtering} 
\label{sec:robust} 

Since early-stopping helps when $\alpha$ is small, we can use FL-compatible filtering techniques to reduce the effective $\alpha$ at any round. 
Filtering algorithms are designed to separate malicious data from benign data, and they often use techniques from robust statistics and outlier detection. 
Many proposed defenses can be viewed as filtering algorithms, including (Multi-)Krum,  activation clustering , and SPECTRE  (more complete list in Appendix \ref{app:related}).
Any of these can be adapted to the FL setting to reduce the effective $\alpha$.

\subsection{Shadow Learning: A Defense Framework}
\label{sec:algo}

For simplicity, we first explain the framework assuming that the defender knows the target class $\ell$; we then explain how to relax this assumption (pseudocode in Algorithm \ref{alg:framework_extend}). 

\paragraph{Training.}
 Shadow learning combines the prior insights by training two models: the \emph{backbone model} and the \emph{shadow model} (Framework~\ref{alg:early}). 
The backbone model is trained to be {\em backdoored} but stable (i.e., insensitive to  distribution shifts and  filtering by our algorithm). 
In training, it is updated in each round using all (including malicious) client data. 
At test time, we only use it if the prediction is \emph{not} the target label. 
Otherwise, we resort to the shadow model,
which is trained on filtered data that removes suspected malicious clients at each round.
This filtering reduces the effective $\alpha$ for the shadow model. 
Finally, the shadow model is early-stopped to avoid backdoor leakage.

\begin{figure}
\centering
\vspace{-0.3cm}
\includegraphics[width=0.6\linewidth]{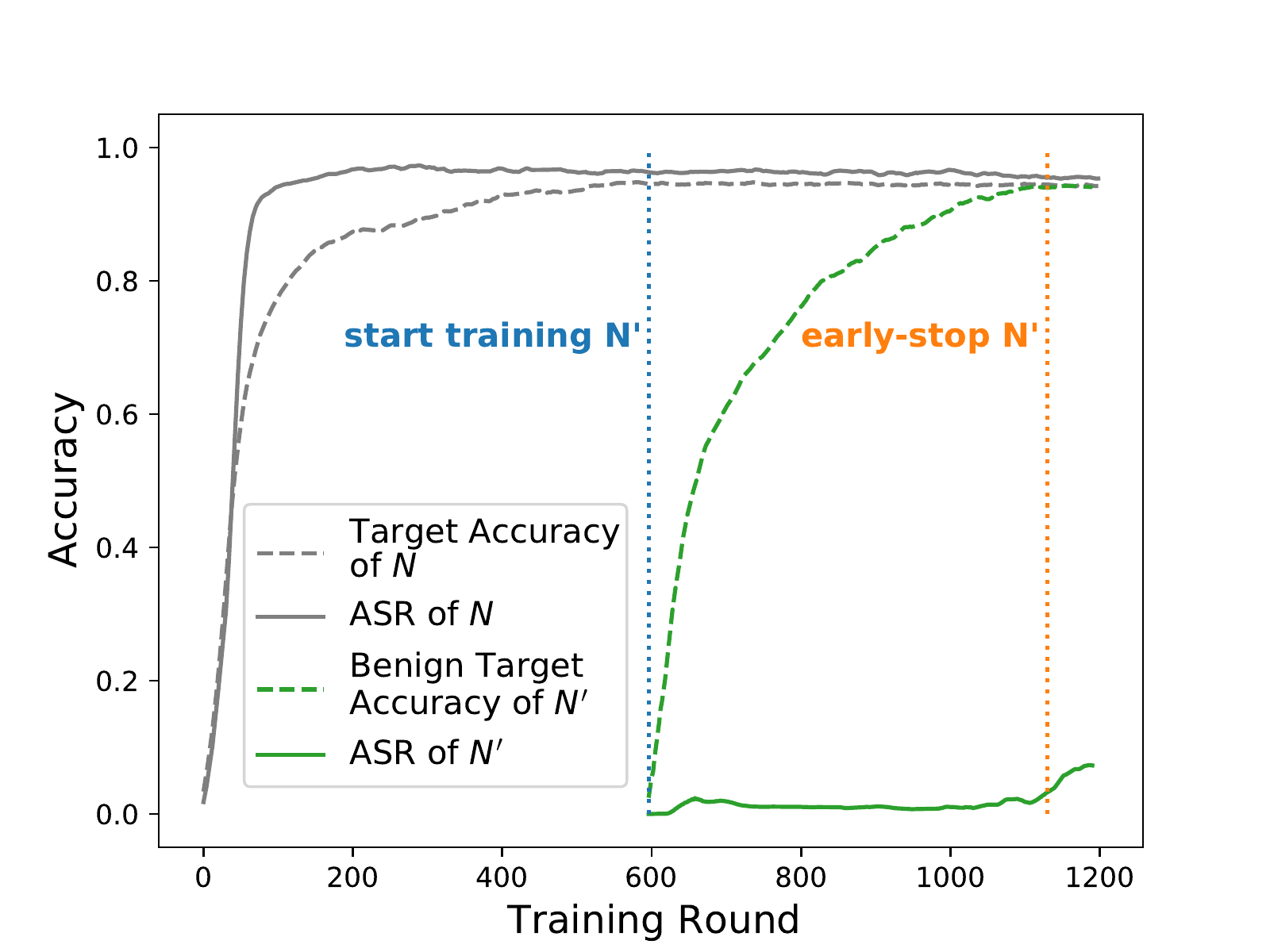}
\caption{Example of training dynamics of the backbone model $\backbone$ and shadow model $\shadow$}
\label{fig:shadow-training}
\vspace{-0.2cm}
\end{figure}

In more detail,
the \emph{backbone model} $\backbone$ is trained continually on all available client data, and thus is backdoored (see Figure \ref{fig:shadow-training}). However, it still serves two purposes: $(i)$ it is used to predict the non-target classes (since it has seen all the past training data and hence is resilient to changing datasets), and $(ii)$ it is used to learn parameters of a filter that can filter out poisoned model updates. 
Concretely, in each training round $r$, the backbone model $\backbone$ is updated on client data from all the clients in the  set $\clientset_{r}$ (line~\ref{line:backbone} Alg.~\ref{alg:framework}). 


In parallel, we train the \emph{shadow model} $\shadow$ on filtered model updates and also early stopped such that it can robustly make predictions for the target label. As it is early stopped, we need to retrain the model periodically from a random initialization in every \emph{retraining period} to handle changing data distributions. A retraining period starts every time the target label distribution changes; in practice, we choose rounds where the training accuracy of the backbone model $\backbone$ on samples with the target label $\targetlabel$ changes by more than a threshold $\epsilon_1$ (lines \ref{line:renew1}-\ref{line:renew2}). 
At the start of each retraining period, the server randomly initializes a new shadow model  $\shadow$ of the same architecture as $\backbone$. 
We say  $\backbone$ has converged on $\targetlabel$ if the difference between the training accuracy on samples with label $\targetlabel$ between two consecutive rounds is smaller than a convergence threshold $\epsilon_2$. 
The next time backbone $\backbone$ converges on the target label $\targetlabel$ distribution (call it round $r_0$, line \ref{line:converge}), we commence training $\shadow$ on filtered client sets $\{\mathcal{C}'_{r}\}_{r\geq r_0}$ until it is early-stopped. 

Concretely, consider retraining a shadow model from round $r_0$ (e.g., in Figure \ref{fig:shadow-training}, $r_0=600$).
In each round $r \geq r_0$, we recompute the filtered client set $\mathcal{C}_{r}'$ whose data is used to train the shadow model $\shadow$. 
This is done by first having each client $c\in \clientset_{r}$ locally average the representations\footnote{E.g., these can be taken from the penultimate layer of $\backbone$.} of samples in $\database_c$ with target label $\targetlabel$; this average is sent to the server for filtering. 
To get the filter, in the first collection round (i.e., $r=600$), the server calls {\sc GetThreshold}, which returns filter parameters $\boldsymbol \theta$ and a threshold $T$; these are used in {\sc Filter} (Alg. \ref{alg:filter}) to remove malicious clients. 

Although our framework supports any filtering algorithm, we empirically find SPECTRE \cite{spectre} to be an effective filtering algorithm (e.g., it has the slowest backdoor leakage in Fig. \ref{fig:backdoor_leakage}) and use it for the remainder of this work. If the filtering algorithm is SPECTRE, {\sc GetThreshold} returns the parameters $\boldsymbol \theta = (\hat {\boldsymbol \Sigma},\hat{\boldsymbol \mu},T,{\boldsymbol U})$ for the robust covariance $\hat{ \boldsymbol  \Sigma}$, 
robust mean $\hat{\boldsymbol \mu}$, filtering threshold $T$, and an orthonormal matrix $U$ representing the top $k$ PCA vectors for the representations $\{{\boldsymbol h}_j\in{\mathbb R}^d\}_{j\in\mathcal{C}_{r}}$. 
The filtering threshold $T$ is set in this case as the $1.5\bar{\alpha}|\mathcal{C}|$-th largest QUE score, which intuitively captures how ``abnormal" a sample is (Alg. \ref{alg:que}), where $\bar{\alpha}$ is an estimated upper bound on the malicious rate.
The SPECTRE filtering process is detailed in Algorithms \ref{alg:threshold} and \ref{alg:filter} (Appendix \ref{sce:alg}).

The shadow network $\shadow$ is early-stopped once the training accuracy on benign target samples converges. To determine the early-stopping point,  clients send the training accuracy on samples with label $\targetlabel$ to the server. If the difference between the average values of the largest $(1-\bar{\alpha})$-fraction training accuracy in two consecutive rounds is smaller than  $\epsilon_2$, $\shadow$ is early-stopped.

We illustrate the training process of the backbone model $\backbone$ and shadow model $\shadow$ in one retraining period with malicious rate $\alpha = 0.15$ under CIFAR-10  in Figure \ref{fig:shadow-training}.
Once the backbone model $\backbone$ converges on the target label $\targetlabel$ (illustrated by the blue dotted line), the server starts training the shadow model $\shadow$ based on the filtered client set. $\shadow$ is early-stopped once its training accuracy on benign target samples converges (illustrated by the orange dotted line).

\paragraph{Testing.}
At test time, all unlabeled samples are first passed through backbone $\backbone$. If predicted as the target label $\targetlabel$, the sample is passed through the early-stopped shadow network $\shadow$, whose prediction is taken as the final output.

\begin{algorithm}[htbp] 
    \label{alg:early}
    \LinesNumbered
	\BlankLine
	\SetKwInOut{Input}{input}
	\SetKwInOut{Output}{output}
	\caption{Shadow learning framework}
	\label{alg:framework}
	\Input{malicious rate upper bound $\bar{\alpha}$, target label $\targetlabel$,
	retraining threshold $\epsilon_1$, convergence threshold $\epsilon_2$, dimension $k$, filtering hyperparameters $\boldsymbol \beta$.}
	\BlankLine
	Initialize the networks $\backbone$, $\shadow$; \  $converge \leftarrow \text{False}$; \  $filter\_learned \leftarrow \text{False}$\;
	\For{\text{each training round} $r$}{
	Train the backbone network $\backbone$ with client set $\mathcal{C}_r$\label{line:backbone}\;
	The server collects training accuracy on samples with label $\targetlabel$ of each client in $\mathcal{C}_r$, and calculates the mean value $A_N^{(r)}$\;
	\If{$converge$  \ {\rm {and}} \ $\left|A_N^{(r)}-A_{N}^{(r-1)}\right|>\epsilon_1$ \label{line:renew1}}{ 
	    $converge \leftarrow \text{False}$; \ 
	    $filter\_learned \leftarrow \text{False}$\;
	    Initialize the shadow network $\shadow$\label{line:renew2}\;
	}
	{\bf if} {\em not} $converge$ \ {\rm {and}} \ $\left|A_N^{(r)}-A_{N}^{(r-1)}\right|<\epsilon_2$
	 {\bf \ then } $converge \leftarrow \text{True}$ \label{line:converge}\; 
	\If{$converge$}{
		Each client $j$ in $\mathcal{C}_r$ uploads the averaged representation $\boldsymbol{h}_j^{(r)}$ of samples with label $\targetlabel$\;
		\If{not $filter\_learned$\label{line:threshold1}}{
			$\boldsymbol{\theta}, T \leftarrow \text{{\sc GetThreshold}}\left(\left\{\boldsymbol{h}_j^{(r)}\right\}_{j\in \mathcal{C}_r}, \bar{\alpha}, k, \boldsymbol \beta \right)$ {\hfill [Algorithm \ref{alg:threshold}]}
			
			$filter\_learned \leftarrow \text{True}$; \ 
			$early\_stop \leftarrow \text{False}$\label{line:threshold2}\;
			}
		\If{not $early\_stop$}{
			$\mathcal{C}'_r\leftarrow \text{{\sc Filter}}\left(\left\{\boldsymbol{h}_j^{(r)}\right\}_{j\in \mathcal{C}_r}, \boldsymbol{\theta}, T, \boldsymbol \beta \right)$ 
			\label{line:filter}
			{\hfill [Algorithm \ref{alg:filter}]}
			
			Train $\shadow$ with client set $\mathcal{C}'_r$\;
			The server collects training accuracy on samples with label $\targetlabel$ of each client in $\mathcal{C}'_r$, and calculates the mean of the largest $(1-\bar{\alpha})$-fraction values as $A_{N'}^{(r)}$\;
			{\bf if } 
			$\left|A_{N'}^{(r)}-A_{N'}^{(r-1)}\right|<\epsilon_2$ {\bf \  then } 
				$early\_stop \leftarrow \text{True}$.
			
			}
		}
	}
\end{algorithm}


\paragraph{Unknown target label $\ell$.}
Suppose that instead of knowing $\ell$ exactly, the defender knows it to be in some set $S_\ell\subseteq [L]$. 
In this case, our framework generalizes by learning a different shadow network $N'_{y}$ and filter for each label $y \in S_\ell$.
It then chooses a set of labels $S_\ell' \subseteq S_\ell$ whose filters have the greatest separation between estimated benign and malicious clients. 
For instance, under SPECTRE, this is done by comparing QUE scores for estimated outliers compared to inliers for each label (Alg. \ref{alg:framework_extend}, App. \ref{algorithm_extended}).
At test time, samples are first passed through the backbone $\backbone$. If the label prediction $y$ is in the filtered target set $S'_{\targetlabel}$, the sample is passed through the early-stopped shadow network $\shadow_y$, whose prediction is taken as the final output.

\section{Analysis}
\label{sec:analysis}

Using SPECTRE as our filtering algorithm, we can theoretically analyze a simplified setting of shadow learning.  
This analysis includes the first theoretical justification of defenses based on robust covariance estimation and outlier detection (e.g., \cite{spectre}), including in non-FL settings.
Specifically, assuming clean and poisoned representations are drawn i.i.d. from different Gaussian distributions, we show that  {\sc GetThreshold} (Alg.~\ref{alg:threshold}) and {\sc Filter} (Alg.~\ref{alg:filter})  reduce  the number of corrupt clients polynomially in $\alpha$ (Theorem~\ref{thm:main}). 
Using predictions from the early-stopped shadow network, this guarantees a correct prediction  (Corollary~\ref{coro:early}).

\begin{asmp} 
    \label{asmp:main} 
      We assume that the representation of the clean and poisoned data points that have the target label are i.i.d.~samples from  $d$-dimensional Gaussian distributions ${\cal N}(\mu_c,\Sigma_c)$ and ${\cal N}(\mu_p,\Sigma_p)$, respectively. 
      The combined representations are  i.i.d.~samples from a mixture distribution $ (1-\alpha){\cal N}(\mu_c,\Sigma_c) + \alpha {\cal N}(\mu_p,\Sigma_p)$ known as Huber contamination   \cite{huber1992robust}. 
      We assume that $\|\Sigma_c^{-1/2}\Sigma_p \Sigma_c^{-1/2}\| \leq \xi < 1 $, where $\|\cdot \|$is the spectral norm. 
\end{asmp} 

The separation, $\Delta=\mu_p-\mu_c$, between the clean and the corrupt  points plays a significant role, especially the magnitude  $\rho = \| \Sigma_c^{-1/2}\Delta\|$. 
We show that {\sc GetThreshold} and {\sc Filter} significantly reduce the poisoned fraction $\alpha$, as long as the separation $\rho$ is sufficiently  larger than (a function of) the poison variance $\xi$.
In the following, $n_r \triangleq |\clientset_r|$ denotes the number of clients in a round $r$. 

\begin{thm}[Utility guarantee for {\sc Threshold} and {\sc Filter}] \label{thm:main} 
For any $m \in{\mathbb Z}_+$ and a large enough $n_r=\Omega((d^2/\alpha^3){\rm polylog}(d/\alpha))$ and small enough $\alpha=O(1)$, under Assumption~\ref{asmp:main}, there exist  positive constants $c_m,c_m'>0$ that only depend on the target exponent $m>0$ such that if the separation is large enough, $\rho  \geq  c_m \sqrt{\log(1/\alpha) + \xi}$, 
then the fraction of the poisoned data clients after {\sc GetThreshold} in Algorithm~\ref{alg:threshold} and {\sc Filter} in Algorithm~\ref{alg:filter} is bounded by 
    $
        \frac{| S_{\rm poison} \setminus S_{\rm filter} |}{n_r-|S_{\rm filter}|} \;\leq\; c'_m \alpha^{m}
    $, 
    with probability $9/10$
    where $S_{\rm filter}$ is the set of client updates that did not pass the {\sc Filter} and $S_{\rm poison}$ is the set of poisoned client updates. 
\end{thm} 
The proof (Appendix~\ref{sec:main_proof}) connects recent results on high-dimensional robust covariance estimation \cite{diakonikolas2017being} to the classical results of \cite{davis1970rotation} to argue that the top eigenvector of the empirical covariance matrix is aligned with the direction of the poisoned representations; this enables effective filtering. 
Theorem \ref{thm:main} suggests that we can select $m=3$ to reduce the fraction of corrupted clients from $\alpha$ to $c'_m \alpha^3$, as long as the separation between clean and poisoned data is large enough. The main tradeoff is in the condition $\rho/\sqrt{\log(1/\alpha)+\xi} \geq c_m$, where the LHS can be interpreted as the Signal-to-Noise  Ratio (SNR) of the problem of detecting poisoned updates. If the SNR is large, the detection succeeds. 

The next result shows that one can get a clean model by early-stopping a model trained on such a filtered dataset. 
This follows as a corollary of \cite[Theorem 2.2]{li2020gradient}; 
it critically relies on an assumption on a $(\varepsilon_0,M)$-clusterable dataset  $\{(x_i\in{\mathbb R}^{d_0} , y_i\in{\mathbb R})\}_{i=1}^{n_r}$  and  overparametrized two-layer neural network models defined in Assumption~\ref{asmp:cluster} in the Appendix~\ref{sec:coro_asmp}. 
If the fraction of malicious clients $\alpha$  is sufficiently small,  early stopping  prevents a two-layer neural network from learning the backdoor. This suggests that our approach can defend against backdoor attacks; {\sc GetThreshold} and {\sc Filter} effectively reduce the fraction of corrupted data, which strengthens our early stopping defense. We further discuss the necessity of robust covariance estimation in Appendix~\ref{app:robust}.

\begin{coro}[{Utility guarantee for early stopping; corollary of \cite[Theorem 2.2]{li2020gradient}}]
    \label{coro:early}
    Under the ($\alpha,n_r,\varepsilon_0,\varepsilon_1,M,\totallabels,\hat M,C,\lambda,W$)-model in Assumption~\ref{asmp:cluster}, 
    starting with  $W_0 \in{\mathbb R}^{\hat M\times d_0}$ with i.i.d.~${\cal N}(0,1)$ entries, there exists $c>0$ such that if 
    $
        \alpha  \leq   {1}/{4(\totallabels-1)} 
    $, 
    where $\totallabels$ is the number of classes,  
    then $\tau=c \|C\|^2/\lambda$ steps of gradient descent on the loss $\loss(W; x,y)=(1/2)(f_W(x)-y)^2 $ for a two-layer neural network parametrized by $W\in{\mathbb R}^{\hat M \times d_0}$ with learning rate $\eta=cM/(n_r \|C\|^2)$ outputs  $W_T$ that correctly predicts the clean label of  a test data ${\boldsymbol x}_{\rm test}$ regardless of whether it is poisoned or not 
    with probability $1- 3/M-Me^{-100d_0}$.  
\end{coro}

\section{Experimental results}
\label{sec:experiments}

We evaluate shadow learning (with R-SPECTRE filtering) on 4 datasets: EMNIST \cite{emnist}, CIFAR-10, 
CIFAR-100 \cite{cifar10}, and Tiny ImageNet \cite{le2015tiny} datasets.
We compare with 8 state-of-the-art defense algorithms under the federated setting: 
 SPECTRE (gradient- and representation-based))\cite{spectre}, Robust Federated Aggregation (RFA) \cite{geomedian}, Norm Clipping, Noise Adding \cite{backdoor}, CRFL \cite{crfl}, FLAME \cite{flame}, Multi-Krum \cite{krum} and FoolsGold \cite{foolsgold}.
We give a detailed explanation of each in Appendix~\ref{sec:compete}, with hyperparameter settings. 

We consider a worst-case adversary that corrupts a full $\alpha$ fraction of clients in each round. In Appendix \ref{app:adaptive_alpha}, we study a relaxed threat model where the adversary can adapt their corrupt client fraction $\alpha(t)$ over time to take advantage of shadow model learning. 

\subsection{Defense under Homogeneous and Static Clients}
\label{section:homogeneous}

In the homogeneous EMNIST dataset, we shuffle the dataset and 100 images are distributed  to each client. 
We assume the target label $\ell=1$
 is known to the defender; we show what happens for unknown $\ell$ in Sec. \ref{app:unknown_label}. 
 We train the model for $1200$ rounds to model continuous training.
 Recall from Figure~\ref{fig:backdoor_leakage} that existing defenses  suffer from backdoor leakage when $\alpha$ is small ($\alpha=0.03$). 
 In contrast,  Algorithm~\ref{alg:framework} achieves an ASR of $0.0013$ for EMNIST and $0.0092$ for CIFAR-10.
Algorithm \ref{alg:framework} also offers strong protection at higher $\alpha$.
For $\alpha$ as high as 0.45, Table~\ref{table:homo_EMNIST} shows that Algorithm~\ref{alg:framework} has an ASR below 0.06 and MTA above $0.995$ on EMNIST, whereas existing defenses all suffer from backdoor leakage. 


\begin{table}[htbp]
\caption{ASR for EMNIST under continuous training shows the advantage of Algorithm~\ref{alg:framework}, while others suffer  backdoor leakage.}
\label{table:homo_EMNIST}
\small
\centering
\vspace{4mm}
\begin{tabular}{ccccc}
\toprule
Defense $\setminus$
$\alpha$ & 0.15 & 0.25 & 0.35 & 0.45
\\\midrule
Noise Adding & 1.00 & 1.00 & 1.00 & 1.00\vspace{0.8mm}\\
\vspace{0.8mm}
\makecell[c]{Clipping and\\ Noise Adding} & 1.00 & 1.00 & 1.00 & 1.00\\
RFA & 1.00 & 1.00 & 1.00 & 1.00\\
Multi-Krum & 1.00 & 1.00 & 1.00 & 1.00\\
FoolsGold & 0.9972 & 1.00 & 1.00 & 1.00\\
FLAME & 1.00 & 1.00 & 1.00 & 1.00\\
CRFL & 0.9873 & 0.9892 & 0.9903 & 0.9884\\
G-SPECTRE & 0.9948 & 1.00 & 1.00 & 1.00\\
R-SPECTRE & 0.9899 & 0.9934 & 1.00 & 1.00
\vspace{0.8mm}\\
\makecell[c]{Shadow Learning\\ 
(label-level /\\ user-level)} & \makecell[c]{\textbf{0.0067}\ / \\ 0.0216} & \makecell[c]{\textbf{0.0101}\ /\\ 0.0367} & \makecell[c]{\textbf{0.0312}\ /\\ 0.0769} & \makecell[c]{\textbf{0.0502}\ /\\ 0.1338}\\
\bottomrule
\end{tabular}
\end{table}

In some scenarios, it might be infeasible to get access to average representations for the target label. In this case,  we propose a user-level defense whose ASR is also shown in the bottom row. We observe that ASR degrades gracefully, when switching to the user-level defense,  
a variant of our algorithm in which each user uploads the averaged representation over all samples rather than samples with the target label. Data homogeneity enables Algorithm \ref{alg:framework} to distinguish malicious clients only based on averaged representations over all samples, without knowing the target label.
We experiment with heterogeneous clients in Section~\ref{section:heterogeneous}. 


\begin{table}[htbp]
\caption{ASR for CIFAR-10 under continuous training 
shows the advantage of Algorithm~\ref{alg:framework}, while others suffer  backdoor leakage.}
\label{table:cifar10}
\small
\centering
\vspace{4mm}
\begin{tabular}{ccccc}
\toprule
Defense $\setminus$
$\alpha$ & 0.15 & 0.25 & 0.35 & 0.45
\\\midrule
Noise Adding & 0.9212 & 0.9177 & 0.9388 & 0.9207
\vspace{0.8mm}\\
\vspace{0.8mm}
\makecell[c]{Clipping and\\ Noise Adding} & 0.9142 & 0.9247 & 0.9282 & 0.9338\\
RFA & 0.9353 & 0.9528 & 0.9563 & 0.9598\\
Multi-Krum & 0.9254 & 0.9196 & 0.9219 & 0.9301\\
FoolsGold & 0.8969 & 0.9038 & 0.9157 & 0.9143\\
FLAME & 0.9328 & 0.9267 & 0.9291 & 0.9331\\
CRFL & 0.8844 & 0.8731 & 0.8903 & 0.8891\\
G-SPECTRE & 0.8754 & 0.9001 & 0.9142 & 0.9193\\
R-SPECTRE & 0.7575 & 0.8826 & 0.8932 & 0.9091
\vspace{0.8mm}\\
\makecell[c]{Shadow Learning
} & \makecell[c]{\textbf{0.0140}} & \makecell[c]{\textbf{0.0355}} & \makecell[c]{\textbf{0.0972}} & \makecell[c]{\textbf{0.1865}}\\
\bottomrule
\end{tabular}
\end{table}

Compared to EMNIST, CIFAR-10 is more difficult to learn. The smaller learning rate results in less effective early stopping  and thus leads to slightly larger ASR in Table \ref{table:cifar10}. 
We show the ASR for CIFAR-100 and Tiny-ImageNet in Appendix \ref{app:homo_more}.



\subsection{Ablation study under distribution drift}
\label{section:time_varying}

We next provide an ablation study to illustrate why the components of shadow learning are all necessary. 
We run this evaluation under a distribution drift scenario,
where the distribution changes every $400$ rounds. 
We use homogeneous EMNIST in the first phase $e_1$, then reduce the number of samples with labels $2$-$5$ to $10\%$ of the original  in phase $e_2$, 
i.e., we reduce the number of images with the label in $\{2, 3, 4, 5\}$ from $10$ to $1$ for each client. 
Again, the backdoor target label is $1$. 

\begin{figure}
\vspace{-0.3cm}
\centering
\includegraphics[width=0.6\linewidth]{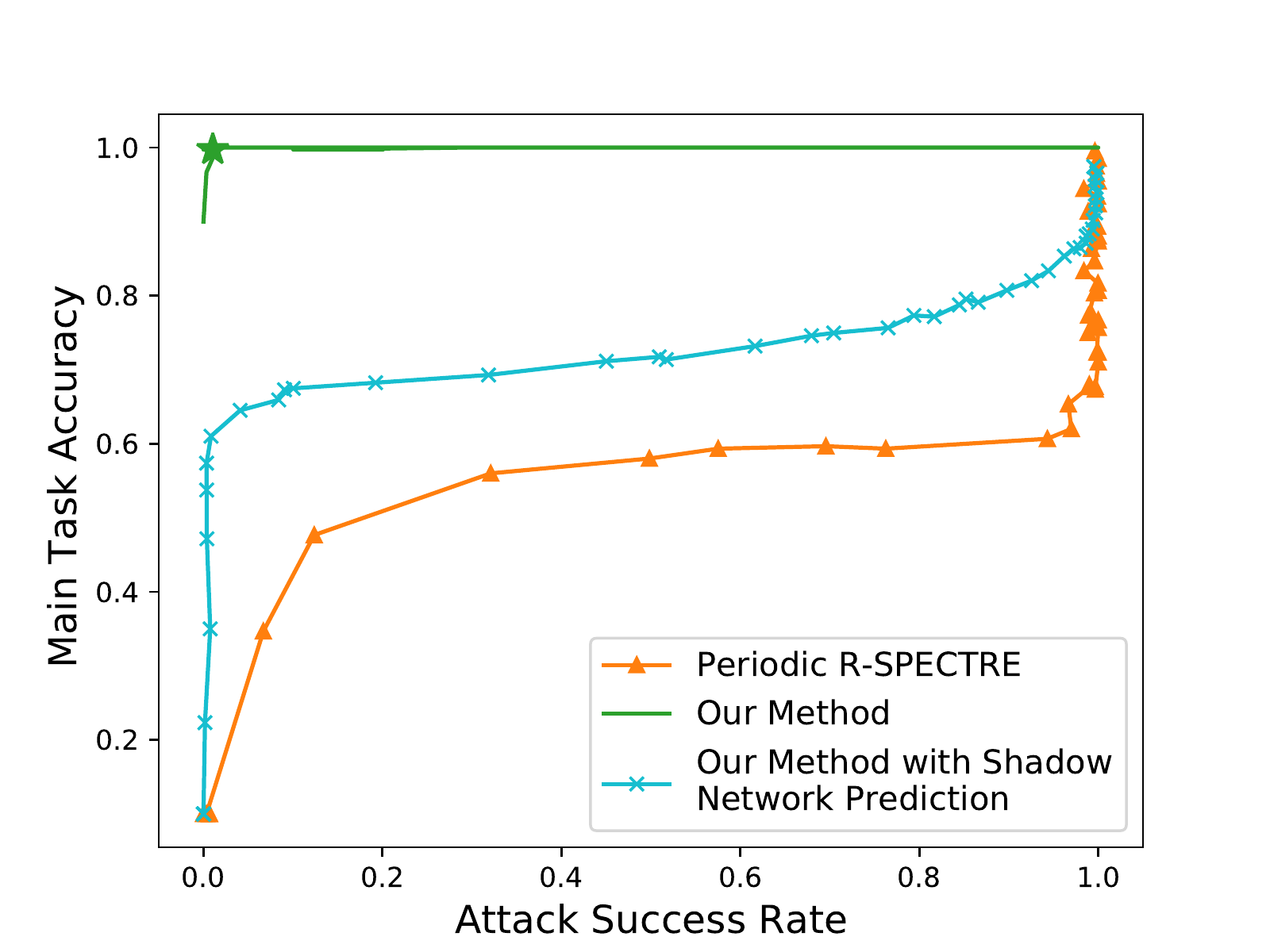}
\caption{Ablation study for shadow learning under distribution drift. Simplified variants of our framework (e.g., periodic R-SPECTRE, shadow network prediction) significantly degrade the MTA-ASR Tradeoff with $\alpha = 0.15$, as achieving top-left is ideal. 
}
\label{fig:time_varying}
\vspace{-0.15cm}
\end{figure}

We compare shadow learning with two simpler variants: Periodic R-SPECTRE and Shadow Network Prediction. 
In periodic R-SPECTRE, the backbone $N$ is retrained from scratch with SPECTRE every $R=400$ rounds; filtering is done in every round.
In Shadow Network Prediction, we use the same training as our framework, but prediction only uses the shadow network (as opposed to using the backbone network as in Algorithm~\ref{alg:framework}). 
The MTA-ASR tradeoffs of all three algorithms in the second phase  $e_2$ are shown in Fig.~\ref{fig:time_varying}.
Results for different periods and types of data distribution drift are shown in Appendix \ref{app:dynamic}.

Fig.\ref{fig:time_varying} shows that for $\alpha=0.15$, 
full shadow learning achieves an MTA-ASR point of $(0.9972, 0.0103)$ (green star),
but the simplified variants perform significantly worse. 
Shadow network prediction does better than periodic R-SPECTRE, as it conducts client filtering only after the convergence of backbone network $\backbone$ on label $\targetlabel$,
while the latter filters in every round.
This indicates that filtering after $\backbone$ converges gives better  performance.
However, shadow network prediction suffers from the lack of training samples with labels $2$-$5$, leading to poor performance.

\begin{table}[htbp]
\caption{ASR under EMNIST partitioned as the original data}
\label{table:EMNIST}
\centering
\vspace{4mm}
\small
\begin{tabular}{ccccc}
\toprule
Defense $\setminus$
$\alpha$ & 0.15 & 0.25 & 0.35 & 0.45
\\\midrule
\makecell[c]{Label-level RFA} & 1.00 & 1.00 & 1.00 & 1.00\\
\makecell[c]{R-SPECTRE} & 0.9967 & 1.00 & 1.00 & 1.00
\vspace{0.8mm}\\
\vspace{0.8mm}
\makecell[c]{Shadow Learning\\ (Sample-level)} & \textbf{0.0107} & \textbf{0.0166} & \textbf{0.0301} & \textbf{0.0534}\\
Shadow Learning & 0.0307 & 0.0378 & 0.0836 & 0.1106
\vspace{0.8mm}\\
\vspace{0.8mm}
\makecell[c]{Shadow Learning\\ (User-level)} & 0.5719 & 0.5912 & 0.6755 & 0.7359\\
\bottomrule
\end{tabular}
\end{table}

\subsection{Defense without knowing target label $\ell$}
\label{app:unknown_label}

We next evaluate shadow learning when the defender only knows that the target label falls into a target set $S_\targetlabel$ on EMNIST.
In our generalized framework (Algorithm \ref{alg:framework_extend}), we set the training round threshold as $R=50$, and set the filtering ratio as $\kappa = 0.2$.
To analyze the averaged ASR and target label detection success rate, we set $\alpha = 0.3$ and vary the size of $S_\targetlabel$ from $2$ to $10$ (i.e., at $10$, the defender knows nothing about $\ell$). 
For each experimental setting, we run our framework $20$ times. 

\begin{figure}[htbp]
\centering
\includegraphics[width=0.6\linewidth]{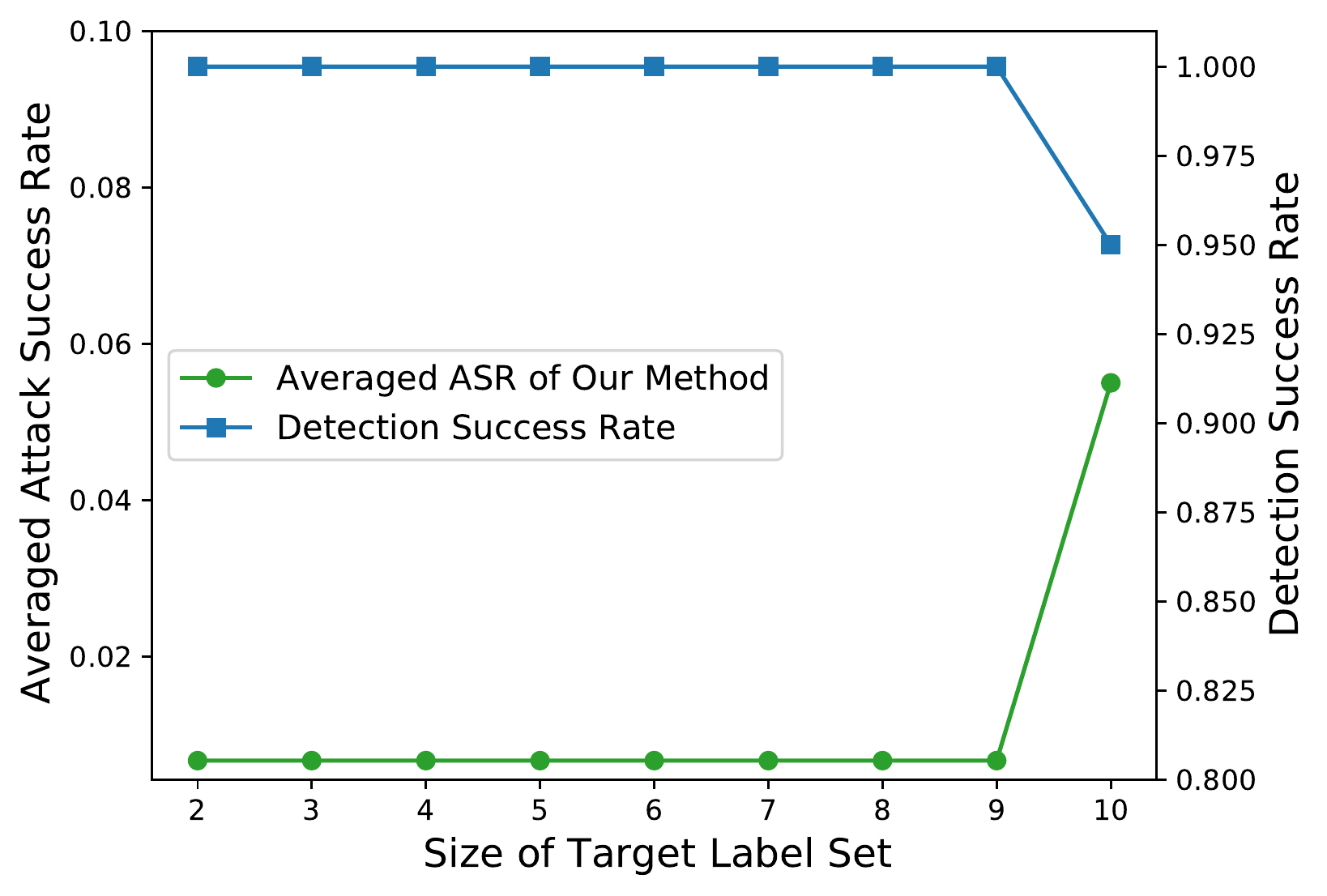}
\caption{Defense Without Knowing the Exact Target Label with $\alpha = 0.3$}
\label{fig:adaptive_target_label}
\end{figure}


As observed in Figure \ref{fig:adaptive_target_label}, when $|S_\targetlabel|\leq 9$, the averaged ASR of our method is near $0$, and the target label can be detected successfully all the time, i.e., target label $\targetlabel$ is in the filter target set $S'_{\targetlabel}$. When $|S_\targetlabel|= 10$, i.e., when the defender has no information about the target label, the averaged ASR and the detection success rates degrade gracefully (notice the different scales for the two vertical axes on Figure \ref{fig:adaptive_target_label}).
This shows that \textbf{even without knowledge of the target label, shadow learning is still effective}. 

\subsection{Defense under Client Heterogeneity}
\label{section:heterogeneous}

We finally evaluate the ASR of Algorithm~\ref{alg:framework} and its two variants:  sample-level and user-level defenses. 
In the sample level version, each user uploads the representations of samples with the target label without averaging, and the samples regarded as backdoor images are filtered out. This weakens privacy but is more robust against heterogeneous clients as shown in Table~\ref{table:EMNIST}. On the other hand, user-level defense fails as heterogeneity of clients makes it challenging to detect corrupt users from user-level aggregate statistics.
The label-level RFA and R-SPECTRE  are evaluated after the network is  trained for $1200$ rounds. More experimental results are provided in Appendix~\ref{app:hetero}.

\subsection{Other settings}
More experimental results including other datasets, defending against adaptive attack strategies (\ref{app:adaptive_alpha} and \ref{sec:fl_attack}), different trigger designs (\ref{sec:trigger}), and robustness to differentially-private noise \ref{sec:noise}, are shown in Appendix~\ref{sec:additional_exp}.






\section{Conclusion}
\label{conclusion}
Motivated by the successes of filtering-based defenses against backdoor attack in the non-FL setting, we propose a novel framework for defending against backdoor attacks in the federated continual learning setting. The main idea is to use such filters to reduce the fraction of corrupt model updates significantly, and then train an early stopped model (called a shadow model) on those filtered updates  to get a clean model.  This combination of filtering and early-stopping significantly improves upon existing defenses and we provide a theoretical justification of our approach. One of the main technical innovations is the parallel training of the backbone and the shadow models, which is critical for obtaining a reliable filter (via the backbone model) and a trustworthy predictor (via the shadow model). Experimenting on four vision datasets and comparing against eight baselines, we show significant improvement on defense against backdoor attacks in FL.

\section*{Acknowledgement}
This material is based upon work supported by the National Science Foundation Convergence Accelerator grant CA-2040675, the U.S. Army Research Office and the U.S. Army Futures Command under Contract No. W911NF-20-D-0002, 
NSF grants CNS-2002664, IIS-1929955, DMS-2134012, CCF-2019844 as a part of NSF Institute for Foundations of Machine Learning (IFML), and  CNS-2112471 as a part of NSF AI Institute for Future Edge Networks and Distributed Intelligence (AI-EDGE). 
The content of the information does not necessarily reflect the position or the policy of the government and no official endorsement should be inferred.
This work was also supported in part by gift grants from J.P. Morgan Chase, Bosch, Siemens, Google, and Intel.


\begin{thebibliography}{10}

\bibitem{awan2021contra}
Sana Awan, Bo~Luo, and Fengjun Li.
\newblock Contra: Defending against poisoning attacks in federated learning.
\newblock In {\em European Symposium on Research in Computer Security}, pages
  455--475. Springer, 2021.

\bibitem{bagdasaryan2020backdoor}
Eugene Bagdasaryan, Andreas Veit, Yiqing Hua, Deborah Estrin, and Vitaly
  Shmatikov.
\newblock How to backdoor federated learning.
\newblock In {\em International Conference on Artificial Intelligence and
  Statistics}, pages 2938--2948. PMLR, 2020.

\bibitem{barni2019new}
Mauro Barni, Kassem Kallas, and Benedetta Tondi.
\newblock A new backdoor attack in cnns by training set corruption without
  label poisoning.
\newblock In {\em 2019 IEEE International Conference on Image Processing
  (ICIP)}, pages 101--105. IEEE, 2019.

\bibitem{biggio2014poisoning}
Battista Biggio, Konrad Rieck, Davide Ariu, Christian Wressnegger, Igino
  Corona, Giorgio Giacinto, and Fabio Roli.
\newblock Poisoning behavioral malware clustering.
\newblock In {\em Proceedings of the 2014 workshop on artificial intelligent
  and security workshop}, pages 27--36, 2014.

\bibitem{blanchard2017machine}
Peva Blanchard, El~Mahdi El~Mhamdi, Rachid Guerraoui, and Julien Stainer.
\newblock Machine learning with adversaries: Byzantine tolerant gradient
  descent.
\newblock {\em Advances in Neural Information Processing Systems}, 30, 2017.

\bibitem{krum}
Peva Blanchard, El~Mahdi El~Mhamdi, Rachid Guerraoui, and Julien Stainer.
\newblock Machine learning with adversaries: Byzantine tolerant gradient
  descent.
\newblock {\em Advances in Neural Information Processing Systems}, 30, 2017.

\bibitem{chai2022one}
Shuwen Chai and Jinghui Chen.
\newblock One-shot neural backdoor erasing via adversarial weight masking.
\newblock {\em arXiv preprint arXiv:2207.04497}, 2022.

\bibitem{chen2018detecting}
Bryant Chen, Wilka Carvalho, Nathalie Baracaldo, Heiko Ludwig, Benjamin
  Edwards, Taesung Lee, Ian Molloy, and Biplav Srivastava.
\newblock Detecting backdoor attacks on deep neural networks by activation
  clustering.
\newblock {\em arXiv preprint arXiv:1811.03728}, 2018.

\bibitem{chen2019deepinspect}
Huili Chen, Cheng Fu, Jishen Zhao, and Farinaz Koushanfar.
\newblock Deepinspect: A black-box trojan detection and mitigation framework
  for deep neural networks.
\newblock In {\em IJCAI}, volume~2, page~8, 2019.

\bibitem{chen2022quarantine}
Tianlong Chen, Zhenyu Zhang, Yihua Zhang, Shiyu Chang, Sijia Liu, and Zhangyang
  Wang.
\newblock Quarantine: Sparsity can uncover the trojan attack trigger for free.
\newblock In {\em Proceedings of the IEEE/CVF Conference on Computer Vision and
  Pattern Recognition}, pages 598--609, 2022.

\bibitem{cheneffective}
Weixin Chen, Baoyuan Wu, and Haoqian Wang.
\newblock Effective backdoor defense by exploiting sensitivity of poisoned
  samples.
\newblock In {\em Advances in Neural Information Processing Systems}, 2022.

\bibitem{chen2017targeted}
Xinyun Chen, Chang Liu, Bo~Li, Kimberly Lu, and Dawn Song.
\newblock Targeted backdoor attacks on deep learning systems using data
  poisoning.
\newblock {\em arXiv preprint arXiv:1712.05526}, 2017.

\bibitem{cheng2021deep}
Siyuan Cheng, Yingqi Liu, Shiqing Ma, and Xiangyu Zhang.
\newblock Deep feature space trojan attack of neural networks by controlled
  detoxification.
\newblock In {\em Proceedings of the AAAI Conference on Artificial
  Intelligence}, volume~35, pages 1148--1156, 2021.

\bibitem{chou2018sentinet}
Edward Chou, Florian Tram{\`e}r, Giancarlo Pellegrino, and Dan Boneh.
\newblock Sentinet: Detecting physical attacks against deep learning systems.
\newblock {\em arXiv preprint arXiv:1812.00292}, 2018.

\bibitem{emnist}
Gregory Cohen, Saeed Afshar, Jonathan Tapson, and Andre Van~Schaik.
\newblock Emnist: Extending mnist to handwritten letters.
\newblock In {\em 2017 international joint conference on neural networks
  (IJCNN)}, pages 2921--2926. IEEE, 2017.

\bibitem{davis1970rotation}
Chandler Davis and William~Morton Kahan.
\newblock The rotation of eigenvectors by a perturbation. iii.
\newblock {\em SIAM Journal on Numerical Analysis}, 7(1):1--46, 1970.

\bibitem{diakonikolas2019robust}
Ilias Diakonikolas, Gautam Kamath, Daniel Kane, Jerry Li, Ankur Moitra, and
  Alistair Stewart.
\newblock Robust estimators in high-dimensions without the computational
  intractability.
\newblock {\em SIAM Journal on Computing}, 48(2):742--864, 2019.

\bibitem{diakonikolas2017being}
Ilias Diakonikolas, Gautam Kamath, Daniel~M Kane, Jerry Li, Ankur Moitra, and
  Alistair Stewart.
\newblock Being robust (in high dimensions) can be practical.
\newblock In {\em International Conference on Machine Learning}, pages
  999--1008. PMLR, 2017.

\bibitem{do2022towards}
Kien Do, Haripriya Harikumar, Hung Le, Dung Nguyen, Truyen Tran, Santu Rana,
  Dang Nguyen, Willy Susilo, and Svetha Venkatesh.
\newblock Towards effective and robust neural trojan defenses via input
  filtering.
\newblock In {\em Computer Vision--ECCV 2022: 17th European Conference, Tel
  Aviv, Israel, October 23--27, 2022, Proceedings, Part V}, pages 283--300.
  Springer, 2022.

\bibitem{feng2022stealthy}
Le~Feng, Sheng Li, Zhenxing Qian, and Xinpeng Zhang.
\newblock Stealthy backdoor attack with adversarial training.
\newblock In {\em ICASSP 2022-2022 IEEE International Conference on Acoustics,
  Speech and Signal Processing (ICASSP)}, pages 2969--2973. IEEE, 2022.

\bibitem{foolsgold}
Clement Fung, Chris~JM Yoon, and Ivan Beschastnikh.
\newblock Mitigating sybils in federated learning poisoning.
\newblock {\em arXiv preprint arXiv:1808.04866}, 2018.

\bibitem{gu2017badnets}
Tianyu Gu, Brendan Dolan-Gavitt, and Siddharth Garg.
\newblock Badnets: Identifying vulnerabilities in the machine learning model
  supply chain.
\newblock {\em arXiv preprint arXiv:1708.06733}, 2017.

\bibitem{guan2022few}
Jiyang Guan, Zhuozhuo Tu, Ran He, and Dacheng Tao.
\newblock Few-shot backdoor defense using shapley estimation.
\newblock In {\em Proceedings of the IEEE/CVF Conference on Computer Vision and
  Pattern Recognition}, pages 13358--13367, 2022.

\bibitem{guo2021aeva}
Junfeng Guo, Ang Li, and Cong Liu.
\newblock Aeva: Black-box backdoor detection using adversarial extreme value
  analysis.
\newblock {\em arXiv preprint arXiv:2110.14880}, 2021.

\bibitem{guo2023scale}
Junfeng Guo, Yiming Li, Xun Chen, Hanqing Guo, Lichao Sun, and Cong Liu.
\newblock Scale-up: An efficient black-box input-level backdoor detection via
  analyzing scaled prediction consistency.
\newblock In {\em ICLR}, 2023.

\bibitem{harikumar2022defense}
Haripriya Harikumar, Santu Rana, Kien Do, Sunil Gupta, Wei Zong, Willy Susilo,
  and Svetha Venkastesh.
\newblock Defense against multi-target trojan attacks.
\newblock {\em arXiv preprint arXiv:2207.03895}, 2022.

\bibitem{spectre}
Jonathan Hayase, Weihao Kong, Raghav Somani, and Sewoong Oh.
\newblock {SPECTRE}: Defending against backdoor attacks using robust
  statistics.
\newblock In {\em International Conference on Machine Learning}, pages
  4129--4139. PMLR, 2021.

\bibitem{hayase2022few}
Jonathan Hayase and Sewoong Oh.
\newblock Few-shot backdoor attacks via neural tangent kernels.
\newblock {\em arXiv preprint arXiv:2210.05929}, 2022.

\bibitem{he2016deep}
Kaiming He, Xiangyu Zhang, Shaoqing Ren, and Jian Sun.
\newblock Deep residual learning for image recognition.
\newblock In {\em Proceedings of the IEEE conference on computer vision and
  pattern recognition}, pages 770--778, 2016.

\bibitem{hu2021trigger}
Xiaoling Hu, Xiao Lin, Michael Cogswell, Yi~Yao, Susmit Jha, and Chao Chen.
\newblock Trigger hunting with a topological prior for trojan detection.
\newblock {\em arXiv preprint arXiv:2110.08335}, 2021.

\bibitem{huang2019neuroninspect}
Xijie Huang, Moustafa Alzantot, and Mani Srivastava.
\newblock Neuroninspect: Detecting backdoors in neural networks via output
  explanations.
\newblock {\em arXiv preprint arXiv:1911.07399}, 2019.

\bibitem{huber1992robust}
Peter~J Huber.
\newblock Robust estimation of a location parameter.
\newblock In {\em Breakthroughs in statistics}, pages 492--518. Springer, 1992.

\bibitem{ilyas2019adversarial}
Andrew Ilyas, Shibani Santurkar, Dimitris Tsipras, Logan Engstrom, Brandon
  Tran, and Aleksander Madry.
\newblock Adversarial examples are not bugs, they are features.
\newblock In {\em Advances in Neural Information Processing Systems}, pages
  125--136, 2019.

\bibitem{jagielski2020auditing}
Matthew Jagielski, Jonathan Ullman, and Alina Oprea.
\newblock Auditing differentially private machine learning: How private is
  private sgd?
\newblock {\em Advances in Neural Information Processing Systems},
  33:22205--22216, 2020.

\bibitem{konevcny2016federated}
Jakub Kone{\v{c}}n{\`y}, H~Brendan McMahan, Daniel Ramage, and Peter
  Richt{\'a}rik.
\newblock Federated optimization: Distributed machine learning for on-device
  intelligence.
\newblock {\em arXiv preprint arXiv:1610.02527}, 2016.

\bibitem{cifar10}
Alex Krizhevsky, Geoffrey Hinton, et~al.
\newblock Learning multiple layers of features from tiny images.
\newblock 2009.

\bibitem{laskov2014practical}
Pavel Laskov.
\newblock Practical evasion of a learning-based classifier: A case study.
\newblock In {\em 2014 IEEE symposium on security and privacy}, pages 197--211.
  IEEE, 2014.

\bibitem{le2015tiny}
Ya~Le and Xuan Yang.
\newblock Tiny imagenet visual recognition challenge.
\newblock {\em CS 231N}, 7(7):3, 2015.

\bibitem{lee2018simple}
Kimin Lee, Kibok Lee, Honglak Lee, and Jinwoo Shin.
\newblock A simple unified framework for detecting out-of-distribution samples
  and adversarial attacks.
\newblock In {\em Advances in Neural Information Processing Systems}, pages
  7167--7177, 2018.

\bibitem{li2020gradient}
Mingchen Li, Mahdi Soltanolkotabi, and Samet Oymak.
\newblock Gradient descent with early stopping is provably robust to label
  noise for overparameterized neural networks.
\newblock In {\em International conference on artificial intelligence and
  statistics}, pages 4313--4324. PMLR, 2020.

\bibitem{li2019invisible}
Shaofeng Li, Minhui Xue, Benjamin Zi~Hao Zhao, Haojin Zhu, and Xinpeng Zhang.
\newblock Invisible backdoor attacks on deep neural networks via steganography
  and regularization.
\newblock {\em IEEE Transactions on Dependable and Secure Computing},
  18(5):2088--2105, 2020.

\bibitem{li2021anti}
Yige Li, Xixiang Lyu, Nodens Koren, Lingjuan Lyu, Bo~Li, and Xingjun Ma.
\newblock Anti-backdoor learning: Training clean models on poisoned data.
\newblock {\em Advances in Neural Information Processing Systems}, 34, 2021.

\bibitem{li2021neural}
Yige Li, Xixiang Lyu, Nodens Koren, Lingjuan Lyu, Bo~Li, and Xingjun Ma.
\newblock Neural attention distillation: Erasing backdoor triggers from deep
  neural networks.
\newblock In {\em International Conference on Learning Representations}, 2021.

\bibitem{li2022untargeted}
Yiming Li, Yang Bai, Yong Jiang, Yong Yang, Shu-Tao Xia, and Bo~Li.
\newblock Untargeted backdoor watermark: Towards harmless and stealthy dataset
  copyright protection.
\newblock {\em arXiv preprint arXiv:2210.00875}, 2022.

\bibitem{li2022backdoor}
Yiming Li, Yong Jiang, Zhifeng Li, and Shu-Tao Xia.
\newblock Backdoor learning: A survey.
\newblock {\em IEEE Transactions on Neural Networks and Learning Systems},
  2022.

\bibitem{li2020open}
Yiming Li, Ziqi Zhang, Jiawang Bai, Baoyuan Wu, Yong Jiang, and Shu-Tao Xia.
\newblock Open-sourced dataset protection via backdoor watermarking.
\newblock {\em arXiv preprint arXiv:2010.05821}, 2020.

\bibitem{li2022black}
Yiming Li, Mingyan Zhu, Xue Yang, Yong Jiang, and Shu-Tao Xia.
\newblock Black-box ownership verification for dataset protection via backdoor
  watermarking.
\newblock {\em arXiv preprint arXiv:2209.06015}, 2022.

\bibitem{liang2017enhancing}
Shiyu Liang, Yixuan Li, and Rayadurgam Srikant.
\newblock Enhancing the reliability of out-of-distribution image detection in
  neural networks.
\newblock {\em arXiv preprint arXiv:1706.02690}, 2017.

\bibitem{liu2021rvfr}
Jing Liu, Chulin Xie, Krishnaram Kenthapadi, Oluwasanmi~O Koyejo, and Bo~Li.
\newblock Rvfr: Robust vertical federated learning via feature subspace
  recovery.
\newblock 2021.

\bibitem{liu2018fine}
Kang Liu, Brendan Dolan-Gavitt, and Siddharth Garg.
\newblock Fine-pruning: Defending against backdooring attacks on deep neural
  networks.
\newblock In {\em International Symposium on Research in Attacks, Intrusions,
  and Defenses}, pages 273--294. Springer, 2018.

\bibitem{liu2017trojaning}
Yingqi Liu, Shiqing Ma, Yousra Aafer, Wen-Chuan Lee, Juan Zhai, Weihang Wang,
  and Xiangyu Zhang.
\newblock Trojaning attack on neural networks.
\newblock 2017.

\bibitem{liu2020reflection}
Yunfei Liu, Xingjun Ma, James Bailey, and Feng Lu.
\newblock Reflection backdoor: A natural backdoor attack on deep neural
  networks.
\newblock In {\em European Conference on Computer Vision}, pages 182--199.
  Springer, 2020.

\bibitem{ma2022beatrix}
Wanlun Ma, Derui Wang, Ruoxi Sun, Minhui Xue, Sheng Wen, and Yang Xiang.
\newblock The" beatrix''resurrections: Robust backdoor detection via gram
  matrices.
\newblock {\em arXiv preprint arXiv:2209.11715}, 2022.

\bibitem{madry2017towards}
Aleksander Madry, Aleksandar Makelov, Ludwig Schmidt, Dimitris Tsipras, and
  Adrian Vladu.
\newblock Towards deep learning models resistant to adversarial attacks.
\newblock {\em arXiv preprint arXiv:1706.06083}, 2017.

\bibitem{flame}
Thien~Duc Nguyen, Phillip Rieger, Huili Chen, Hossein Yalame, Helen
  M{\"o}llering, Hossein Fereidooni, Samuel Marchal, Markus Miettinen, Azalia
  Mirhoseini, Shaza Zeitouni, et~al.
\newblock $\{$FLAME$\}$: Taming backdoors in federated learning.
\newblock In {\em 31st USENIX Security Symposium (USENIX Security 22)}, pages
  1415--1432, 2022.

\bibitem{phan2022invisible}
Huy Phan, Yi~Xie, Jian Liu, Yingying Chen, and Bo~Yuan.
\newblock Invisible and efficient backdoor attacks for compressed deep neural
  networks.
\newblock In {\em ICASSP 2022-2022 IEEE International Conference on Acoustics,
  Speech and Signal Processing (ICASSP)}, pages 96--100. IEEE, 2022.

\bibitem{geomedian}
Krishna Pillutla, Sham~M Kakade, and Zaid Harchaoui.
\newblock Robust aggregation for federated learning.
\newblock {\em arXiv preprint arXiv:1912.13445}, 2019.

\bibitem{qi2023revisiting}
Xiangyu Qi, Tinghao Xie, Yiming Li, Saeed Mahloujifar, and Prateek Mittal.
\newblock Revisiting the assumption of latent separability for backdoor
  defenses.
\newblock In {\em ICLR}, 2023.

\bibitem{qi2022circumventing}
Xiangyu Qi, Tinghao Xie, Saeed Mahloujifar, and Prateek Mittal.
\newblock Circumventing backdoor defenses that are based on latent
  separability.
\newblock {\em arXiv preprint arXiv:2205.13613}, 2022.

\bibitem{savazzi2021opportunities}
Stefano Savazzi, Monica Nicoli, Mehdi Bennis, Sanaz Kianoush, and Luca
  Barbieri.
\newblock Opportunities of federated learning in connected, cooperative, and
  automated industrial systems.
\newblock {\em IEEE Communications Magazine}, 59(2):16--21, 2021.

\bibitem{shokri2020bypassing}
Reza Shokri et~al.
\newblock Bypassing backdoor detection algorithms in deep learning.
\newblock In {\em 2020 IEEE European Symposium on Security and Privacy
  (EuroS\&P)}, pages 175--183. IEEE, 2020.

\bibitem{steinhardt2017certified}
Jacob Steinhardt, Pang Wei~W Koh, and Percy~S Liang.
\newblock Certified defenses for data poisoning attacks.
\newblock In {\em Advances in neural information processing systems}, pages
  3517--3529, 2017.

\bibitem{backdoor}
Ziteng Sun, Peter Kairouz, Ananda~Theertha Suresh, and H~Brendan McMahan.
\newblock Can you really backdoor federated learning?
\newblock {\em arXiv preprint arXiv:1911.07963}, 2019.

\bibitem{tang2021demon}
Di~Tang, XiaoFeng Wang, Haixu Tang, and Kehuan Zhang.
\newblock Demon in the variant: Statistical analysis of dnns for robust
  backdoor contamination detection.
\newblock In {\em USENIX Security Symposium}, pages 1541--1558, 2021.

\bibitem{tao2022better}
Guanhong Tao, Guangyu Shen, Yingqi Liu, Shengwei An, Qiuling Xu, Shiqing Ma,
  Pan Li, and Xiangyu Zhang.
\newblock Better trigger inversion optimization in backdoor scanning.
\newblock In {\em Proceedings of the IEEE/CVF Conference on Computer Vision and
  Pattern Recognition}, pages 13368--13378, 2022.

\bibitem{tolpegin2020data}
Vale Tolpegin, Stacey Truex, Mehmet~Emre Gursoy, and Ling Liu.
\newblock Data poisoning attacks against federated learning systems.
\newblock In {\em European Symposium on Research in Computer Security}, pages
  480--501. Springer, 2020.

\bibitem{tran2018spectral}
Brandon Tran, Jerry Li, and Aleksander Madry.
\newblock Spectral signatures in backdoor attacks.
\newblock In {\em Advances in Neural Information Processing Systems}, pages
  8000--8010, 2018.

\bibitem{wang2020certifying}
Binghui Wang, Xiaoyu Cao, and Neil~Zhenqiang Gong.
\newblock On certifying robustness against backdoor attacks via randomized
  smoothing.
\newblock {\em arXiv preprint arXiv:2002.11750}, 2020.

\bibitem{wang2019neural}
Bolun Wang, Yuanshun Yao, Shawn Shan, Huiying Li, Bimal Viswanath, Haitao
  Zheng, and Ben~Y Zhao.
\newblock Neural cleanse: Identifying and mitigating backdoor attacks in neural
  networks.
\newblock In {\em 2019 IEEE Symposium on Security and Privacy (SP)}, pages
  707--723. IEEE, 2019.

\bibitem{wang2020attack}
Hongyi Wang, Kartik Sreenivasan, Shashank Rajput, Harit Vishwakarma, Saurabh
  Agarwal, Jy-yong Sohn, Kangwook Lee, and Dimitris Papailiopoulos.
\newblock Attack of the tails: Yes, you really can backdoor federated learning.
\newblock {\em Advances in Neural Information Processing Systems}, 33, 2020.

\bibitem{wang2022invisible}
Tong Wang, Yuan Yao, Feng Xu, Shengwei An, Hanghang Tong, and Ting Wang.
\newblock An invisible black-box backdoor attack through frequency domain.
\newblock In {\em Computer Vision--ECCV 2022: 17th European Conference, Tel
  Aviv, Israel, October 23--27, 2022, Proceedings, Part XIII}, pages 396--413.
  Springer, 2022.

\bibitem{wang2022confidence}
Tong Wang, Yuan Yao, Feng Xu, Miao Xu, Shengwei An, and Ting Wang.
\newblock Confidence matters: Inspecting backdoors in deep neural networks via
  distribution transfer.
\newblock {\em arXiv preprint arXiv:2208.06592}, 2022.

\bibitem{wang2022adaptive}
Yuhang Wang, Huafeng Shi, Rui Min, Ruijia Wu, Siyuan Liang, Yichao Wu, Ding
  Liang, and Aishan Liu.
\newblock Adaptive perturbation generation for multiple backdoors detection.
\newblock {\em arXiv preprint arXiv:2209.05244}, 2022.

\bibitem{wang2022bppattack}
Zhenting Wang, Juan Zhai, and Shiqing Ma.
\newblock Bppattack: Stealthy and efficient trojan attacks against deep neural
  networks via image quantization and contrastive adversarial learning.
\newblock In {\em Proceedings of the IEEE/CVF Conference on Computer Vision and
  Pattern Recognition}, pages 15074--15084, 2022.

\bibitem{weber2020rab}
Maurice Weber, Xiaojun Xu, Bojan Karlas, Ce~Zhang, and Bo~Li.
\newblock Rab: Provable robustness against backdoor attacks.
\newblock {\em arXiv preprint arXiv:2003.08904}, 2020.

\bibitem{xiang2020revealing}
Zhen Xiang, David~J Miller, Hang Wang, and George Kesidis.
\newblock Revealing perceptible backdoors in dnns, without the training set,
  via the maximum achievable misclassification fraction statistic.
\newblock In {\em 2020 IEEE 30th International Workshop on Machine Learning for
  Signal Processing (MLSP)}, pages 1--6. IEEE, 2020.

\bibitem{crfl}
Chulin Xie, Minghao Chen, Pin-Yu Chen, and Bo~Li.
\newblock Crfl: Certifiably robust federated learning against backdoor attacks.
\newblock In {\em International Conference on Machine Learning}, pages
  11372--11382. PMLR, 2021.

\bibitem{xie2019dba}
Chulin Xie, Keli Huang, Pin-Yu Chen, and Bo~Li.
\newblock Dba: Distributed backdoor attacks against federated learning.
\newblock In {\em International Conference on Learning Representations}, 2019.

\bibitem{yang2017generative}
Chaofei Yang, Qing Wu, Hai Li, and Yiran Chen.
\newblock Generative poisoning attack method against neural networks.
\newblock {\em arXiv preprint arXiv:1703.01340}, 2017.

\bibitem{yao2019latent}
Yuanshun Yao, Huiying Li, Haitao Zheng, and Ben~Y Zhao.
\newblock Latent backdoor attacks on deep neural networks.
\newblock In {\em Proceedings of the 2019 ACM SIGSAC Conference on Computer and
  Communications Security}, pages 2041--2055, 2019.

\bibitem{yue2022model}
Zhihao Yue, Jun Xia, Zhiwei Ling, Ting Wang, Xian Wei, and Mingsong Chen.
\newblock Model-contrastive learning for backdoor defense.
\newblock {\em arXiv preprint arXiv:2205.04411}, 2022.

\bibitem{zeng2021adversarial}
Yi~Zeng, Si~Chen, Won Park, Z~Morley Mao, Ming Jin, and Ruoxi Jia.
\newblock Adversarial unlearning of backdoors via implicit hypergradient.
\newblock {\em arXiv preprint arXiv:2110.03735}, 2021.

\bibitem{zeng2022narcissus}
Yi~Zeng, Minzhou Pan, Hoang~Anh Just, Lingjuan Lyu, Meikang Qiu, and Ruoxi Jia.
\newblock Narcissus: A practical clean-label backdoor attack with limited
  information.
\newblock {\em arXiv preprint arXiv:2204.05255}, 2022.

\bibitem{zhao2022natural}
Feng Zhao, Li~Zhou, Qi~Zhong, Rushi Lan, and Leo~Yu Zhang.
\newblock Natural backdoor attacks on deep neural networks via raindrops.
\newblock {\em Security and Communication Networks}, 2022, 2022.

\bibitem{zhao2022defeat}
Zhendong Zhao, Xiaojun Chen, Yuexin Xuan, Ye~Dong, Dakui Wang, and Kaitai
  Liang.
\newblock Defeat: Deep hidden feature backdoor attacks by imperceptible
  perturbation and latent representation constraints.
\newblock In {\em Proceedings of the IEEE/CVF Conference on Computer Vision and
  Pattern Recognition}, pages 15213--15222, 2022.

\bibitem{zhong2020backdoor}
Haoti Zhong, Cong Liao, Anna~Cinzia Squicciarini, Sencun Zhu, and David Miller.
\newblock Backdoor embedding in convolutional neural network models via
  invisible perturbation.
\newblock In {\em Proceedings of the Tenth ACM Conference on Data and
  Application Security and Privacy}, pages 97--108, 2020.

\bibitem{zhong2022imperceptible}
Nan Zhong, Zhenxing Qian, and Xinpeng Zhang.
\newblock Imperceptible backdoor attack: From input space to feature
  representation.
\newblock {\em arXiv preprint arXiv:2205.03190}, 2022.

\end{thebibliography}


\appendix
\section*{Appendix}


\section{Related Work}
\label{app:related}

For more comprehensive survey of backdoor attacks and defenses, we refer to \cite{li2022backdoor}. 

\paragraph{Backdoor attacks.}
There has been a vast body of attacks on machine learning (ML) pipelines. 
In this work, we consider only \emph{training-time} attacks, in which the attacker modifies either the training data or process to meet their goals, some of which are particularly deteriorating in FL settings \cite{backdoor,wang2020attack}. Inference-time attacks are discussed in several survey papers \cite{madry2017towards,ilyas2019adversarial}.
Further, we do not consider data poisoning attacks, in which the goal of the adversary is simply to decrease the model's prediction accuracy  \cite{yang2017generative, biggio2014poisoning} which has also been studied in the federated setting \cite{backdoor}. 
Instead, we focus on \emph{backdoor attacks}, where the attacker's goal is to train a model to output a target classification label on samples that contain a trigger signal (specified by the attacker) while classifying other samples correctly at test-time \cite{gu2017badnets}.
The most commonly-studied backdoor attack is a \emph{pixel attack}, in which the attacker inserts a small pattern of pixels into a subset of training samples of a given source class, and changes their labels to the target label \cite{gu2017badnets}. Pixel attacks can be effective even when only a small fraction of training data is corrupted \cite{gu2017badnets}.
Many subsequent works have explored other types of backdoor attacks, including (but not limited to) periodic signals \cite{zhong2020backdoor}, feature-space perturbations \cite{chen2017targeted,liu2017trojaning}, reflections \cite{liu2020reflection}, strong trigggers that only require few-shot backdoor examples \cite{hayase2022few}, and FL model replacement that explicitly try to evade anomaly detection \cite{bagdasaryan2020backdoor}.
One productive line of work in in making the trigger human-imperceptible  \cite{li2019invisible,barni2019new,yao2019latent,liu2017trojaning,zhong2022imperceptible,feng2022stealthy,wang2022invisible,wang2022bppattack,phan2022invisible,zhao2022natural}.  
Another line of work attempts to make the trigger stealthy in the latent space \cite{shokri2020bypassing,cheng2021deep,zhao2022defeat,zeng2022narcissus,qi2022circumventing,qi2023revisiting}. However, due to the privacy preserving nature of the federated setting, the attacker is limited in the information on other clients' data in the federated scenario.
Although our evaluation will focus on pixel attacks, our experiments suggest that our insights translate to other types of attacks as well (Appendix \ref{sec:trigger} and \ref{sec:fl_attack}). 
Advances in stronger backdoor attacks led to their use in other related domains, including copyright protection for training data \cite{li2020open,li2022black,li2022untargeted} and auditing differential privacy \cite{jagielski2020auditing}.

\paragraph{Backdoor defenses.}
As mentioned earlier, many backdoor defenses can be viewed as taking one (or more) of three approaches: (1) Malicious data detection, (2) Robust training, and (3) Trigger identification. 

Malicious data detection-based methods exploit the idea that adversarial samples will differ from the benign data distribution (e.g., they may be outliers).   
Many such defenses require access to clean samples \cite{liang2017enhancing,lee2018simple,steinhardt2017certified}, which we assume to be unavailable. 
Others work only when the initial adversarial fraction is large, as in anti-backdoor learning \cite{li2021anti} (see \S~\ref{sec:design}), or small, as in robust vertical FL \cite{liu2021rvfr}; we instead require a method that works across all adversarial fractions. 
Recently, \emph{SPECTRE} proposed using robust covariance estimation to estimate the covariance of the benign data from a (partially-corrupted) dataset \cite{spectre}. 
The data samples are whitened to amplify the spectral signature of corrupted data. 
Malicious samples are then filtered out by thresholding based on a \emph{QUantum Entropy (QUE)} score, which projects the whitened data down to a scalar. 
We use SPECTRE as a building block of our algorithm, and also as a baseline for comparison. 
To adapt SPECTRE to the FL setting, we apply it to gradients (we call this G-SPECTRE) or sample representations (R-SPECTRE), each averaged over a single client's local data with target label $\targetlabel$.
However, we see in \S~\ref{sec:backdoor-leakage} that SPECTRE alone does not work in the continuous learning setting. We propose shadow learning, where a second shadow model is trained on data filtered using one of the malicious data detection-based methods, for example \cite{tran2018spectral,chen2018detecting,spectre}. 
Some other malicious data filtering approaches, such as \cite{huang2019neuroninspect,do2022towards,cheneffective}, require inspecting clients' individual training samples, which is typically not feasible in FL systems.
Instead, 
filtering approaches, such as \cite{ma2022beatrix,tang2021demon,tran2018spectral,chen2018detecting}, that are based on the {\em statistics} of the examples can potentially be used within our shadow model framework.


Robust training methods do not explicitly identify and/or filter outliers; instead, they modify the training (and/or testing) procedure to implicitly remove their contribution. 
For example, \emph{Robust Federated Aggregation (RFA)} provides a robust secure aggregation oracle based on the geometric median \cite{geomedian}.
It is shown to be robust against data poisoning attacks both theoretically and empirically. However, it was not evaluated on backdoor attacks. 
In this work, we adopt RFA as a baseline, and show that it too suffers from backdoor leakage (\S~\ref{sec:backdoor-leakage}).
Other variants of robust training methods require a known bound on the magnitude of the adversarial perturbation; for example, randomized smoothing \cite{wang2020certifying,weber2020rab} ensures that the classifier outputs the same label for all points within a ball centered at a particular sample.  
We assume the radius of adversarial perturbations to be unknown at training time. 
Other approaches again require access to clean data, which we assume to be unavailable. 
Examples include fine-pruning \cite{liu2018fine}, which trains a pruned, fine-tuned model from clean data.

Trigger identification approaches typically examine the training data to infer the structure of a trigger \cite{chen2019deepinspect,chen2022quarantine,tao2022better,guo2021aeva,hu2021trigger, wang2022adaptive, xiang2020revealing, wang2022confidence, chai2022one, harikumar2022defense, yue2022model, guan2022few}. For example, NeuralCleanse \cite{wang2019neural} searches for data perturbations that change the classification of a sample to a target class. SentiNet \cite{chou2018sentinet} uses techniques from model interpretability to identify contiguous, salient regions of input images. These approaches are ill-suited to the FL setting, as they require fine-grained access to training samples and are often tailored to a specific type of backdoor trigger (e.g., pixel attacks).

Finally, note that there is a large body of work defending against data poisoning attacks \cite{backdoor,geomedian,blanchard2017machine,awan2021contra, laskov2014practical,tolpegin2020data}. 
In general, such defenses may not work against backdoor attacks. For example, we show in \S~\ref{sec:backdoor-leakage} and \S~\ref{sec:experiments} that defenses against data poisoning attacks such as RFA \cite{geomedian}, norm clipping \cite{backdoor}, and noise addition \cite{backdoor} are ineffective against backdoor attacks, particularly in the continuous training setting. 

There are other defenses include that are less explored including test-time backdoor defense \cite{guo2023scale} and model unlearning \cite{zeng2021adversarial}.

\section{Algorithm Details}
\label{sce:alg}

In the client filtering process, the collected representations are projected down to a $k$-dimensional space by $\boldsymbol{U}$, then whitened to get $\tilde {\boldsymbol{h}}_j = \boldsymbol{\hat\Sigma}^{-1/2} (\boldsymbol{U}^T\boldsymbol{h}_j- \boldsymbol{\hat\mu} ) \in{\mathbb R}^k$ for all $j\in \mathcal{C}_{r}$. The projection onto ${\boldsymbol U}$ is to reduce computational complexity, which is less critical for the performance of the filter. The whitening with clean covariance $\boldsymbol{\hat\Sigma}$ and clean mean  $\boldsymbol{\hat\mu}$ ensures that the poisoned representations stand out from the clean ones and is critical for the performance of the filter. 
Based on the whitened representations, the server calculates QUE scores (which roughly translates as the scaled-norm of the whitened representation) for all clients and keeps the clients with scores less than the threshold $T$ as $\mathcal{C}'_{r}$ ({\sc Filter} in line \ref{line:filter}).
The details of {\sc GetThreshold} and the client filtering process, {\sc Filter}, are shown in Algorithms \ref{alg:threshold} and \ref{alg:filter}.

\begin{algorithm}[htpb]
	\BlankLine
	\SetKwInOut{Input}{input}
	\SetKwInOut{Output}{output}
	\caption{{\sc GetThreshold} (SPECTRE-based instantiation) }\label{alg:threshold}
	\Input{representation $S = \left\{\boldsymbol{h}_i\right\}_{i=1}^n$, malicious rate upper bound $\bar{\alpha}$, dimension $k$, QUE parameter $\beta$.}
	\BlankLine
	$\boldsymbol{\mu}(S) \leftarrow \frac 1 n \sum_{i=1}^{n} \boldsymbol{h}_i$\;
	$S_1 = \left\{\boldsymbol{h}_i - \boldsymbol{\mu}(S)\right\}_{i=1}^n$\;
	$\boldsymbol{V}, \boldsymbol{\Lambda}, \boldsymbol{U} = \text{SVD}_k(S_1)$\;
	$S_2 \leftarrow \{\boldsymbol{U}^\top \boldsymbol{h}_i\}_{\boldsymbol{h}_i \in S}$\;
	$\boldsymbol{\hat\Sigma}, \boldsymbol{\hat\mu} \leftarrow$ {\sc RobustEst}
	$(S_2, \bar{\alpha})$; \cite{diakonikolas2019robust}
	
	$S_3 \leftarrow \left\{\boldsymbol{\hat\Sigma}^{-1/2}\left(\boldsymbol{\bar{h}}_i-\boldsymbol{\hat\mu}\right)\right\}_{\boldsymbol{\bar{h}}_i \in S_2}$\;
	$\{t_i\}_{i=1}^n\leftarrow \text{\sc QUEscore}(S_3, \beta)$ {\hfill [Algorithm \ref{alg:que}]}
	
	$T \leftarrow \text{the} \  1.5\bar{\alpha}n\text{-th largest value in} \{t_i\}_{i=1}^n$\;
	\Return $\boldsymbol{\hat\Sigma}, \boldsymbol{\hat\mu}, T, \boldsymbol{U}$
\end{algorithm}

\begin{algorithm}[htpb]
	\BlankLine
	\SetKwInOut{Input}{input}
	\SetKwInOut{Output}{output}
	\caption{{\sc Filter} (SPECTRE-based instantiation)}\label{alg:filter}
	\Input{$S = \left\{\boldsymbol{\bar{h}}_i \in \mathbb{R}^k\right\}_{i=1}^n$, estimated covariance $\boldsymbol{\hat\Sigma}$, estimated mean $\boldsymbol{\hat\mu}$, threshold $T$, QUE parameter $\beta$.}
	\BlankLine
	$S' \leftarrow \left\{\boldsymbol{\hat\Sigma}^{-1/2}\left(\boldsymbol{\bar{h}}_i-\boldsymbol{\hat\mu}\right)\right\}_{\boldsymbol{\bar{h}}_i \in S}$\;
	$\{t_i\}\leftarrow \text{\sc QUEscore}(S', \beta)$ {\hfill [Algorithm \ref{alg:que}]}
	
	\Return clients with QUE-scores smaller than $T$
\end{algorithm}

The details of the {\sc QUEscore}\cite{spectre} is shown in Algorithm \ref{alg:que}

\begin{algorithm}[htpb]
	\BlankLine
	\SetKwInOut{Input}{input}
	\SetKwInOut{Output}{output}
	\caption{{\sc QUEscore} \cite{spectre}}\label{alg:que}
	\Input{$S = \left\{\boldsymbol{\tilde{h}}_i \in \mathbb{R}^k\right\}_{i=1}^n$, QUE parameter $\beta$.}
	\BlankLine
	$$
	t_i \leftarrow \frac{\boldsymbol{\tilde{h}}_i^\top Q_\beta \boldsymbol{\tilde{h}}_i}{Tr(Q_\beta)}, \quad \forall i \in [n],
	$$
	
	where 
	$Q_\beta = \text{exp}\left(
	\frac{\beta(\widetilde{\Sigma}-\boldsymbol{\mathrm{I}})}
	{\|\widetilde{\Sigma}\|_2-1}
	\right)$
	and 
	$
	\widetilde{\Sigma} = \frac 1 n \Sigma_{i=1}^n \boldsymbol{\tilde{h}}_i \boldsymbol{\tilde{h}}_i^\top
	$.
	
	\Return $\{t_i\}_{i=1}^n$
	
\end{algorithm}

\section{Shadow Learning Framework Without Knowledge of the Target Label}
\label{algorithm_extended}

For simplicity, we presented Algorithm \ref{alg:framework} using knowledge of the target label $\ell$.
However, this knowledge is not necessary. 
In practice, an adversary may know that the target label falls into a target set $S_\ell \subseteq [L]$. Note that the defender has no information about the target class when $S_\ell = [L]$, i.e., when it contains all labels.
Under this assumption, shadow learning generalizes to  Algorithm \ref{alg:framework_extend}.

At training time, the central server maintains a \emph{backbone model} and multiple \emph{shadow models}, each shadow model corresponding to a potential target label in $S_\targetlabel$.
The backbone model 
$\backbone$ is trained continuously based on Algorithm \ref{alg:framework}. For each shadow model $\shadow_y$, where $y\in S_\targetlabel$, it is first trained for $R$ rounds according to Algorithm \ref{alg:framework}, where each client uploads the averaged representation of samples with label $y$. To distinguish the actual target label, for every shadow network training round $r$, each shadow network $\shadow_y$ calculates the QUE-ratio $\gamma^{\bra{r}}_y = \frac{Q_1}{Q_2}$, where $Q_1$(resp. $Q_2$) is the QUE-score averaged over clients with scores larger (resp. smaller) than $T$, which is the QUE threshold calculated in Algorithm \ref{alg:framework}. 
Compared with non-target labels, the QUE-scores of malicious clients calculated based on the target label $\targetlabel$ are much larger than that of benign clients, and therefore, $\gamma_\targetlabel^{(r)}$ is larger than $\gamma_y^{(r)}$ ($y\in S_\targetlabel\setminus \brc{\targetlabel}$) in most cases.
After the shadow networks have been trained for $R$ rounds, the averaged QUE-ratio $\overline{\gamma}_y$ will be calculated for every $\shadow_y$, according to which the filtered target set $S'_\targetlabel$ can be obtained: $$
S'_{\targetlabel} \leftarrow \brc{y | \overline{\gamma}_y \text{ is among the top }\lceil \kappa |S_\targetlabel| \rceil \text{ largest values of } \brc{\overline{\gamma}_z}_{z\in S_\targetlabel}},
$$ where the hyper-parameter $\kappa$ is the filtering ratio. 
Then only the shadow networks with labels in the filter target set $S'_{\targetlabel}$ will be trained according to Algorithm \ref{alg:framework}.

At test time, all unlabeled samples are first passed through backbone $\backbone$. If the label prediction $y$ falls into the filtered target set $S'_{\targetlabel}$, the sample is passed through the early-stopped shadow network $\shadow_y$, whose prediction is taken as the final output. 
We show that Algorithm~\ref{alg:framework_extend} works well for most $S_\ell$ and degrades gracefully as the set increases in  Figure~\ref{fig:adaptive_target_label}. 

\begin{algorithm}[H]
    \label{alg:early-nolabel}
    \LinesNumbered
	\BlankLine
	\SetKwInOut{Input}{input}
	\SetKwInOut{Output}{output}
	\caption{Shadow learning framework (training) without knowing the exact target label}
	\label{alg:framework_extend}
	\Input{target set $S_\targetlabel$, training round threshold $R$, filtering ratio $\kappa$.}
	\BlankLine
	Initialize the networks $\backbone$, $\brc{\shadow_y}_{y\in S_\targetlabel}$\; $\Gamma_y \leftarrow 0, \ \forall y \in S_\targetlabel$\;
	Train the backbone network $\backbone$ according to Algorithm~\ref{alg:framework}\;
	\For{\text{each shadow network training round} $r$}{
	\If{$r\leq R$}{
	\For{\text{each shadow network $\shadow_y$ where ${y\in S_\targetlabel}$ }}
	{
	Obtain QUE-score for each client and threshold $T$, and train $\shadow_y$ according to Algorithm~\ref{alg:framework}, where each client uploads the averaged representation of samples with label $y$\; 
	$Q_1 \leftarrow \text{QUE-score averaged over clients with scores larger than }T$\;
	$Q_2 \leftarrow \text{QUE-score averaged over clients with scores smaller than }T$\;
	$\gamma^{\bra{r}}_y = \frac{Q_1}{Q_2}$\;
	$\Gamma_y \leftarrow \Gamma_y + \gamma^{\bra{r}}_y$;
	}
	}
	\If{$r = R$}{
	$\overline{\gamma}_y \leftarrow \frac{\Gamma_y}{R},\ \forall y \in S_\targetlabel$\;
	$S'_{\targetlabel} \leftarrow \brc{y | \overline{\gamma}_y \text{ is among the top }\lceil \kappa |S_\targetlabel| \rceil \text{ largest values of } \brc{\overline{\gamma}_z}_{z\in S_\targetlabel}}$\;
	}
	\If{r > R}{
	Train $\shadow_y$, where $y\in S'_\targetlabel$, according to Algorithm \ref{alg:framework}\;
	}
}
\end{algorithm}

\section{Complete Proofs of the Main Theoretical Results} 
\label{Sec:proof} 
We provide proofs of main results and accompanying technical lemmas. We use $c,c',C,C',\ldots$ to denote generic numerical constants that might differ from line to line. 

\subsection{Proof of Theorem~\ref{thm:main}}
\label{sec:main_proof} 

The proof proceeds in two steps. First, we show that under Assumption~\ref{asmp:main}, the direction of the top eigenvector of the empirical covariance is aligned with the direction of the center of the poisoned representations (Lemma~\ref{lem:spectral}). We next show that filtering with the quantum score significantly reduces the number of poisoned clients (Lemma~\ref{lem:quantum}). 

      We let $\hat\Sigma$ be the output of the robust covariance estimator in Algorithm~\ref{alg:threshold}. 
      After whitening by $\hat\Sigma^{-1/2}$, we let 
    $\tilde\Sigma_c = \hat\Sigma^{-1/2}\Sigma_c \hat\Sigma^{-1/2}$,
    $\tilde\Sigma_p = \hat\Sigma^{-1/2}\Sigma_p \hat\Sigma^{-1/2}$, 
    $\tilde \Delta = \hat\Sigma^{-1/2} \Delta$, 
    $\tilde\mu_p = \hat\Sigma^{-1/2}\mu_p$, and 
    $\tilde\mu_c = \hat\Sigma^{-1/2}\mu_c$.
    
The next lemma shows that as we have a larger separation $\|\tilde\Delta\|$, we get a better estimate of the direction of the poisons, $\tilde\Delta/\|\tilde\Delta\|$, using the principal component, $v$, of the whitened representations $S'=\{\Sigma^{-1/2}(h_i - \mu) \}_{i=1}^{n_r}$. The estimation error is measured in $\sin^2$ of the angle between the two. 
    
    \begin{lemma}[Estimating $\tilde\Delta$ with top eigenvector]
    \label{lem:spectral}
    Under the assumptions of Theorem~\ref{thm:main}, 
        \begin{eqnarray}
           \sin^2(v,\tilde\Delta/\|\tilde\Delta\|) \;=\;  1-( v^T \tilde\Delta/\|\tilde\Delta\|  )^2 
            \;\leq \; c  \frac{(\log(1/\alpha) + \xi )^2}{\|\tilde\Delta\|^4 } \;. 
        \end{eqnarray}
    \end{lemma}
    We next show that when projected onto any direction $v$, the number of corrupted client updates passing the filter is determined by how closely aligned $v$ is with the direction of the poisoned data, $\tilde\Delta$, and the magnitude of the separation, $\|\tilde\Delta\|$. 
    
    \begin{lemma}[Quantum score filtering]
        \label{lem:quantum} 
        Under the hypotheses of Theorem~\ref{thm:main}, 
        \begin{eqnarray} 
            \frac{|  S_{\rm poison}\setminus S_{\rm filter}|}{|  (S_{\rm poison} \cup S_{\rm clean})\setminus S_{\rm filter}|} \; \leq \; 
            c' \alpha\, n_r\, Q\Big(\frac{v^T\tilde\Delta - c\sqrt{\log(1/\alpha)}}{\xi^{1/2}} \Big) \;,
        \end{eqnarray} 
        where $Q(t)=\int_{t}^{\infty} (1/\sqrt{2\pi})e^{-\frac{x^2}{2}}dx$ is the tail of the standard Gaussian. 
    \end{lemma}
    Since $ \| \tilde\Delta\|^2 \geq C(\log(1/\alpha) + \xi) $, Lemma~\ref{lem:spectral} implies $v^T\tilde\Delta \geq (1/2) \|\tilde\Delta\|$. Since $ (v^T\tilde\Delta-c\sqrt{\log(1/\alpha)})/\xi^{1/2} \geq C' \sqrt{\log(1/\alpha)/\xi}$, 
    Lemma~\ref{lem:quantum} implies that $|S_{\rm poison} \setminus S_{\rm filter}|\leq \alpha^{C''/\xi} $. Since $\xi<1$, we can make any desired exponent by increasing the separation by a constant factor.

\subsubsection{Proof of Lemma~\ref{lem:spectral}}
\label{sec:spectral_proof}

Let $\tilde\Sigma$ denote the empirical covariance of the whitened representations.  
With a large enough sample size, we have the following bound on the robustly estimated covariance. 
    \begin{thm}[{Robust covariance estimation  \cite{diakonikolas2017being}[Theorem 3.3]}] If the sample size is ${n_r}=\Omega((d^2/\alpha^2){\rm polylog}(d/\alpha))$ with a large enough constant then with probability $9/10$, 
        \label{thm:robustcov}
        \begin{eqnarray}
            \label{eq:robustcov}
            \|\hat\Sigma^{-1/2}\Sigma_c\hat\Sigma^{-1/2} - {\bf I}_d\|_F \;\leq \; c \alpha\log(1/\alpha)\;,
        \end{eqnarray}
        for some universal constant $c>0$ where $\|A\|_F$ denotes the Frobenius norm of a matrix $A$. 
    \end{thm}
Denoting $\tilde\Sigma=\hat\Sigma^{-1/2}  \Sigma_{\rm emp} \hat\Sigma^{-1/2}$ with the empirical covariance of the clean representations denoted by  $\Sigma_{\rm emp}=(1/n)\sum_{i=1}^n (h_i-\mu_{\rm emp})(h_i-\mu_{\rm emp})^T = (1-\alpha)\hat\Sigma_C + \alpha \hat\Sigma_p + \alpha(1-\alpha)\hat\Delta\hat\Delta^T$, where $\hat\Sigma_c$, $\hat\Sigma_p$, and $\hat\Delta$ are the empirical counterparts, we can use this to bound, 
    \begin{align}
        &\|\tilde\Sigma-(1-\alpha){\bf I}_d -\alpha \tilde \Sigma_p - \alpha(1-\alpha)\tilde\Delta\tilde\Delta^T\| 
        \nonumber\\
        &\;\; \leq \;\;
        (1-\alpha)\|\hat\Sigma^{-1/2} (\hat\Sigma_c-\Sigma_c) \hat\Sigma^{-1/2}\| + 
        (1-\alpha)\|\hat\Sigma^{-1/2}\Sigma_c\hat\Sigma^{-1/2} - {\bf I}_d \|  
        \nonumber\\
        & \;\;\;\;
        + \;\;
        \alpha\|\hat\Sigma^{-1/2}(\Sigma_p-\hat\Sigma_p)\hat\Sigma^{-1/2} \| +  \alpha(1-\alpha)\|\hat\Sigma^{-1/2}(\Delta\Delta^T-\hat\Delta\hat\Delta^T )\hat\Sigma^{-1/2} \|
        \nonumber \\ 
        &\;\;\leq  \;\;
        c'\alpha \log(1/\alpha)\;, \label{eq:spectralub1}
    \end{align}
    for a large enough sample size ${n_r}=\tilde \Omega(d^2/\alpha^3)$. Among other things, this implies that 
    \begin{eqnarray}
        \label{eq:spectralLB}
        \|\tilde\Sigma-(1-\alpha){\bf I}\| \;\geq \; \alpha(1-\alpha)\|\tilde\Delta\|^2-c'\alpha\log(1/\alpha)\;.
    \end{eqnarray}
    We use Davis-Kahan theorem to turn a spectral norm bound between two matrices into a angular distance bound between the top singular vectors of the two matrices: 
\begin{eqnarray}
    \sqrt{1- \frac{(v^T\tilde\Delta)^2}{\|\tilde\Delta\|^2}} &\leq & \frac{\| (\|\tilde\Sigma\| -(1-\alpha))vv^T - \alpha(1-\alpha)\tilde\Delta\tilde\Delta^T \|_F}{\|\tilde\Sigma -(1-\alpha){\bf I}\|} \nonumber \\
    &\leq& \frac{\sqrt{2}\| (\|\tilde\Sigma\| -(1-\alpha))vv^T - \alpha(1-\alpha)\tilde\Delta\tilde\Delta^T \|}{\|\tilde\Sigma -(1-\alpha){\bf I}\|} \nonumber \\ 
    &\leq& \frac{2\sqrt{2}\|  \tilde\Sigma -(1-\alpha){\bf I} -  \alpha(1-\alpha)\tilde\Delta\tilde\Delta^T   \|}{\|\tilde\Sigma  -(1-\alpha){\bf I}\|} \nonumber\\ 
    &\leq&  \frac{c\alpha(\log(1/\alpha) + \|\tilde\Sigma_p\|)}{\alpha(1-\alpha)\|\tilde\Delta\|^2-c'\alpha\log(1/\alpha)}\;,
\end{eqnarray}
where $\|A\|$ and $\|A\|_F$ denote spectral and Frobenius norms of a matrix $A$, respectively, the first inequality follows from Davis-Kahan theorem~\cite{davis1970rotation}, the second inequality follows from the fact that Frobenius norm of a rank-2 matrix is bounded by $\sqrt{2}$ times the spectral norm,  and 
the third inequality follows from the fact that $(\|\tilde\Sigma\|-(1-\alpha))vv^T $ is the best rank one approximation of $\tilde\Sigma-(1-\alpha){\bf I}$  and hence $ \|(\|\tilde\Sigma\|-(1-\alpha))vv^T-\tilde\Delta\tilde\Delta^T\| \leq \|(\|\tilde\Sigma\|-(1-\alpha))vv^T-(\tilde\Sigma-(1-\alpha){\bf I})\|+\|(\tilde\Sigma-(1-\alpha){\bf I})-\tilde\Delta\tilde\Delta^T\| \leq 2 \|\tilde\Sigma-(1-\alpha){\bf I}-\tilde\Delta\tilde\Delta^T \|$. The last inequality follows from Eq.~\eqref{eq:spectralub1} 
and  Eq.~\eqref{eq:spectralLB}.
For $\alpha\leq 1/2$ and $\|\tilde\Delta\|\geq \sqrt{\log(1/\alpha)}$, this implies the desired result. 

\subsubsection{Proof of Lemma~\ref{lem:quantum}}
\label{sec:quantum_proof}

We consider a scenario where the QUEscore of a (whitened and centered) representation $\tilde h_i = \hat\Sigma^{-1/2}(h_i-\hat\mu_c)$, where $\hat\mu_c$ is the robust estimate of $\mu_c$, is computed as 
\begin{eqnarray}
    \label{eq:QUEscore}
    \tau^{(\beta )}_i \; = \; 
    \frac{\tilde h_i^TQ_\beta \tilde h_i}{{\rm Tr}(Q_\beta)}
    \;,
\end{eqnarray}
where $Q_\beta=\exp((\alpha/\|\tilde\Sigma\|-1)(\tilde\Sigma-{\bf I}))$.
We analyze the case where we choose  $\beta=\infty$, such that $\tau_i^{(\infty)}=(v^T\tilde h_i)^2 $ and the threshold $T$ returned by SPECTRE satisfies the following. If we have infinite samples and there is no error in the estimates $v$, $\tilde \Sigma$, and $\hat\mu_c$, then 
we have $Q^{-1}((3/4)\alpha) \leq T^{1/2} \leq Q^{-1}((1/4)\alpha)$, which follows from the fact that for the clean data with identity covariance, we can filter out at most $1.5\alpha$ fraction of the data (which happens if we do not filter out any of the poisoned data points) and we can filter out at least  $0.5\alpha$ fraction of the data (which happens if we filter out all the poisoned data). With finite samples and estimation errors in the robust estimates, we get the following: 
\begin{eqnarray} 
    \label{eq:threshold}
    Q^{-1}((3/4)\alpha + \alpha^2/d) - c'\alpha/d \; \leq \; T^{1/2} \; \leq \; Q^{-1}((1/4)\alpha - c'\alpha^2/d)  + c'\alpha/d \;,
\end{eqnarray}
where $Q(\cdot)$ is the tail of a standard Gaussian as defined in Lemma~\ref{lem:quantum} and we used the fact that for a large enough sample size we have $\|v^T\hat\Sigma^{-1/2}(\mu_c-\hat\mu_c)\|\leq c' \alpha/d $.
 
 At test time, when we  filter out data points with QUEscore larger than $T$, 
we have that we filter out at most clean $|S_{\rm clean}\cap S_{\rm filter}| \leq 2\alpha n_r$ representations 
for a large enough $d$. Similarly, we are guaranteed that the remaining poisoned representations are at most  $|S_{\rm poison}\setminus S_{\rm filter}| \leq Q((v^T\tilde\Delta-T^{1/2})/\xi^{1/2}) (\alpha n_r)$. Since from above bound $T^{1/2}\leq c \sqrt{\log(1/ \alpha)} $, this proves the desired bound. 

\subsection{Assumptions for Corollary~\ref{coro:early}} 
\label{sec:coro_asmp}

Corollary~\ref{coro:early} follows as a corollary of \cite{li2020gradient}[Theorem 2.2]. 
This critically relies on an assumption on a $(\varepsilon_0,M)$-clusterable dataset  $\{(x_i\in{\mathbb R}^{d_0} , y_i\in{\mathbb R})\}_{i=1}^{n_r}$  and  overparametrized two-layer neural network models, 
as defined in Assumption~\ref{asmp:cluster} below. 

\begin{asmp}[{($\alpha,n_r,\varepsilon_0,\varepsilon_1,M,\totallabels,\hat M,C,\lambda,W$)-model}]
\label{asmp:cluster} 
 The $(1-\alpha)$ fraction of data points are clean and  originate from $M$ clusters with each cluster containing ${n_r}/M$ data points. Cluster centers are unit norm vectors, $\{\mu_\classindex\in{\mathbb R}^{d_0}\}_{\classindex=1}^M$.  
 An input $x_i$ that belong to the $\classindex$-th cluster obeys $\| {\boldsymbol x}_i - \mu_\classindex\|\leq \varepsilon_0$, with $\varepsilon_0$  denoting the input noise level. The labels $y$ belong to one of the $\totallabels$ classes and we place them evenly in $[-1,1]$ in the training such that labels correspond to $y\in\{-1,-1+1/(\totallabels-1), \ldots,1\}$. 
 We let $C=[\mu_1,\ldots,\mu_M]^T\in {\mathbb R}^{M \times d_0 }$ and define $\lambda=\lambda(C)$ as the minimum eigenvalue of the matrix $CC^T \odot {\mathbb E}[\phi'(Cg)\phi'(Cg)^T] $ for $g\sim{\cal N}(0,{\bf I}_{d_0})$.
 
 All clean data points in the same cluster share the same label. Any two clusters obey $|\mu_\classindex-\mu_{\classindex'}|\geq 2\varepsilon_0+\varepsilon_1$, where $\varepsilon_1$ is the size of the trigger. 
 The corrupted data points are generated from a data point ${\boldsymbol x}_i$ with a source label $y_i=y_{\rm source}$ belonging to one of the clusters for the source. A fixed trigger ${\boldsymbol \delta} $ is added to the input and labelled as a target label $\classindex$ such that the corrupted paired example is $({\bf x}_i+{\boldsymbol \delta},\classindex)$. We train on the combined dataset with $\alpha n_r$ corrupted  points and $(1-\alpha)n_r$ uncorrupted  points.  
 We train a neural network of the form $f_W({\boldsymbol x})=v^T\phi(W {\boldsymbol x})$ for a trainable parameter  $W\in{\mathbb R}^{ \hat M\times d_0}$ and a fixed $v\in{\mathbb R}^{\hat M}$, where the head $v$ is fixed as $1/\sqrt{\hat M}$ for half of the entries and $-1/\sqrt{\hat M}$ for the other half. We assume the activation function $\phi:{\mathbb R}\to{\mathbb R}$ satisfy $ |\phi'(z)|,|\phi''(z)|<c''$ for some constant $c''>0$.
\end{asmp}

\subsection{Necessity of robust covariance estimation} 
\label{app:robust}

To highlight that SPECTRE, and  the robust covariance estimation, is critical in achieving this guarantee, we next show that under the same assumptions, using the QUEscore filtering without whitening fails and also using whitening with non-robust covariance estimation also fails, in the sense that the fraction of the corrupted data is non-decreasing. 
We construct an example within Assumption~\ref{asmp:main} as follows: $\mu_c=0$ and $\Sigma_c=\sigma^2({\bf I} - (1-\delta)uu^T)$ for a unit norm $u$. We place all the poisons at $\mu_p=a u$ with $\xi=0$ and covariance $\Sigma_p$ zero.  We let $\delta\leq a/(c_m^2 \log(1/\alpha))$ such that the separation condition is met. By increasing $\sigma^2$, we can make the inner product of top PCA direction $v_{\rm combined}$ of the combined data and the direction of the poisons $u$ arbitrarily small. Hence, after finding the threshold $T$ in the projected representations $v^T_{\rm combined}h_i$ and using this to filter out the representations, as proposed in \cite{tran2018spectral}, the ratio of the poisons can only increase as all the poisons are placed close to the center of the clean representations after projection. 
The same construction and conclusion holds for the case when we whitened with not the robustly estimated covariance, but the  QUEscore based filter with $\beta=\infty$ projects data first onto the PCA direction of the whitened data, which can be made arbitrarily orthogonal to the direction of the poisons, in high dimensions, i.e. $v_{\rm whitened}^Tu \leq 2/d$ with high probability.
This follows from the fact that after (non-robust) whitening, all directions are equivalent  and the chance of PCA finding the direction of the poisons is uniformly at random over all directions in the $d$ dimensional space. 
    
\section{Experimental Details}
\label{sec:compete}

In the experiments in \S\ref{sec:experiments} and Appendix \ref{sec:additional_exp}, we train on the EMNIST, CIFAR-10, CIFAR-100, and Tiny-ImageNet datasets. 
In the EMNIST dataset, there are 3383 users with roughly 100 images in ten labels per user with heterogeneous label distributions. We train a
convolutional network with two convolutional layers, max-pooling, dropout, and two dense layers. 
The CIFAR-10 dataset has $50,000$ training examples, with $5000$ samples in each label. We  partition those samples to $500$ users uniformly at random and  train a ResNet-18 \cite{he2016deep}.
With $500$ samples in each label, the CIFAR-100 and Tiny-ImageNet datasets have $50,000$ and $100,000$ training examples respectively.
For both datasets, we  partition those samples to $100$ users uniformly at random and  train a ResNet-18.

For all datasets, the server randomly selects $50$ clients each round, and each client trains the current model with the local data with batch size $20$,  learning rate $0.1$, and for two iterations. The server learning rate is $0.5$.  
The attacker tries to make $7$'s predicted as $1$'s for EMNIST, horses as automobiles for CIFAR-10, roses as dolphin for CIFAR-100, and bees as cats for Tiny-ImageNet. The backdoor trigger is a $5$$\times$$5$-pixel black square at the bottom right corner of the image.
An $\alpha$ fraction of the clients are chosen to be malicious, who are given $10$ corrupted samples. 
We set the malicious rate $\alpha$ as its upper bound, i.e., $\alpha = \bar{\alpha}$. 
We study first homogeneous and static settings, and we discuss   heterogeneous and dynamic settings in \S\ref{section:heterogeneous} and \S\ref{section:time_varying}. 
In our framework, we set the retraining threshold $\epsilon_1$ as $2\%$, and the convergence threshold $\epsilon_2$ as $0.05\%$.
We let the dimensionality reduction parameter $k$ be $32$ and set the QUE parameter $\beta$ as $4$ in Algorithm~\ref{alg:framework}.

\paragraph{Baselines.}
SPECTRE \cite{spectre} adopts robust covariance estimation and data whitening to amplify the spectral signature of corrupted data, and then detects backdoor samples based on quantum entropy (QUE)  score.
In federated learning settings, we adopt gradient- and representation-based SPECTRE, which takes as input the gradient updates or sample representations averaged over a single client's local data with target label $\targetlabel$.
For both versions, we conduct the robust estimation and quantum score filtering every training round regardless of the computation constraints.
We let the dimensionality reduction parameter $k$ be $32$ and set the QUE parameter $\beta$ as $4$. 

RFA \cite{geomedian} provides a robust secure aggregation oracle based on the geometric median, which is calculated by the Weiszfeld’s Algorithm. In our experiments, we implement RFA with 4-iteration Weiszfeld’s Algorithm.
We also consider the label-level RFA:
for each label, the  geometric median of the aggregated gradient uploaded from each client is estimated. The server then updates the model based on the averaged geometric medians of all labels.

Norm Clipping defense \cite{backdoor} bounds the norm of each model update to at most some threshold $M$, and Noise Adding method \cite{backdoor} is to add a small amount of Gaussian noise to the updates. In our experiments, we set the norm threshold $M$ as $3$ and add independent Gaussian noise with variance $0.03$ to each coordinate.

Multi-Krum \cite{krum} provides an aggregation rule that only select a subset of uploaded gradients that are close to most of their neighbors. In our experiments, we set the Byzantine parameter $f$ as $50\bar{\alpha}$, where $\bar{\alpha}$ is the malicious rate upper bound, and set the number of selected gradients as $m=20$.

FoolsGold \cite{foolsgold} provides an aggregation method that uses an adaptive learning
rate per client based on inter-client contribution similarity. In our experiments, we set the confidence parameter as $1$.

FLAME \cite{flame} adopts noise adding method, and uses the model clustering and norm clipping approach to reduce the amount of noise added. In our experiments, we set the noise level factors as $\lambda = 0.001$.

CRFL \cite{crfl} trains a certifiably robust FL model by adopting noise adding and norm clipping during the training time and using randomized parameter smoothing during testing. In our experiments, we set the norm threshold $M$ as $3$ and add independent Gaussian noise with variance $0.03$ to each coordinate during training. We use $\sigma_T = 0.01$ to
generate $M = 500$ noisy models in parameter smoothing
procedure, and set the certified radius as $1$ and the error tolerance as $0.001$.

\subsubsection{Resource Costs}
\label{cost}
All algorithms including ours are implemented and performed on a server with two Xeon Processor E5-2680 CPUs. Running all defenses for our experiments took approximately 1000 CPU-core hours.





\section{Additional Experimental Results} 
\label{sec:additional_exp}

\subsection{Backdoor leakage under more datasets}
\label{app:leakage}

\begin{figure}[htbp]
\centering
\subfigure[CIFAR-10]{
\includegraphics[width=0.31\linewidth]{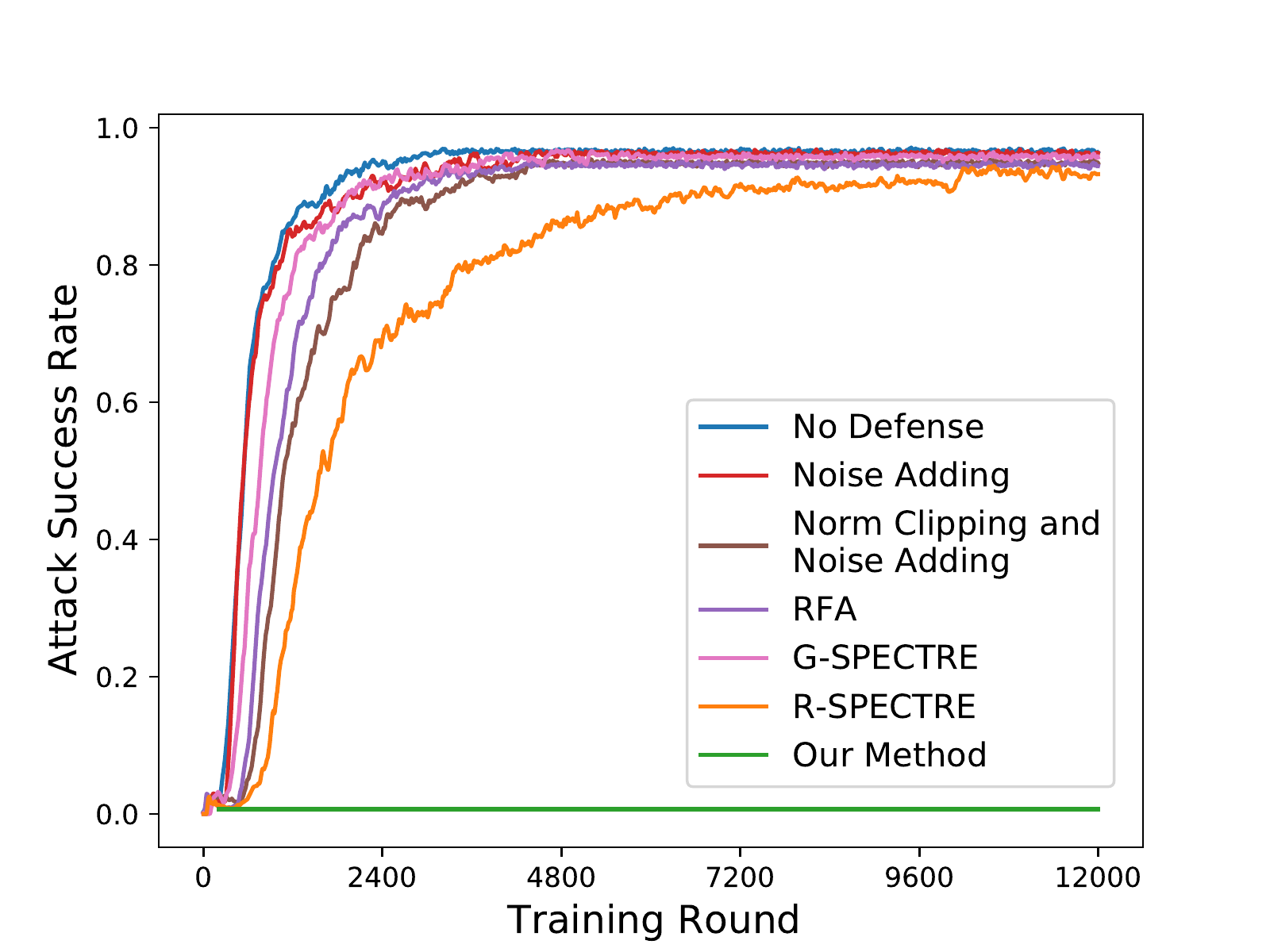}
}
\subfigure[CIFAR-100]{
\includegraphics[width=0.31\linewidth]{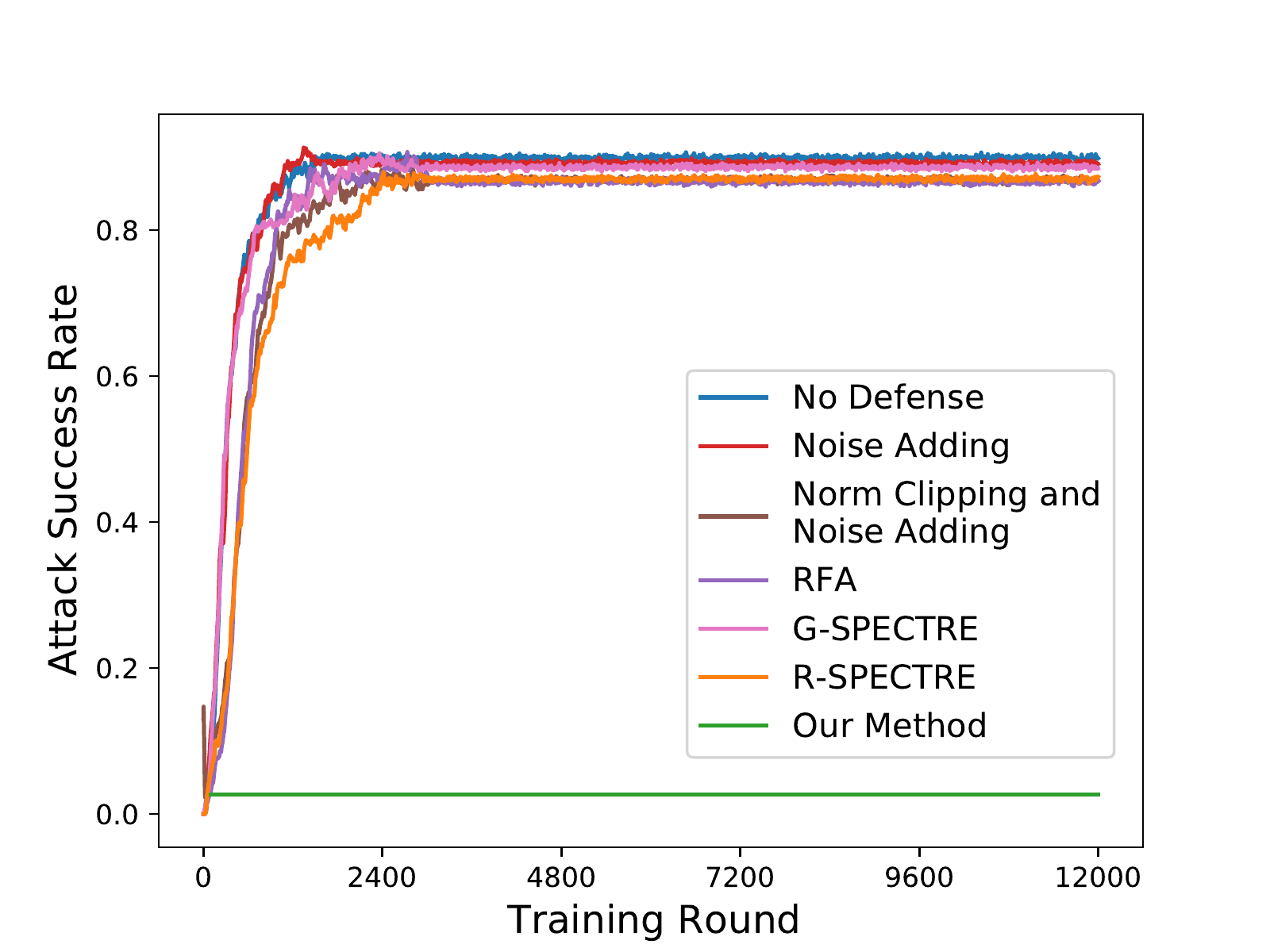}
}
\subfigure[Tiny ImageNet]{
\includegraphics[width=0.31\linewidth]{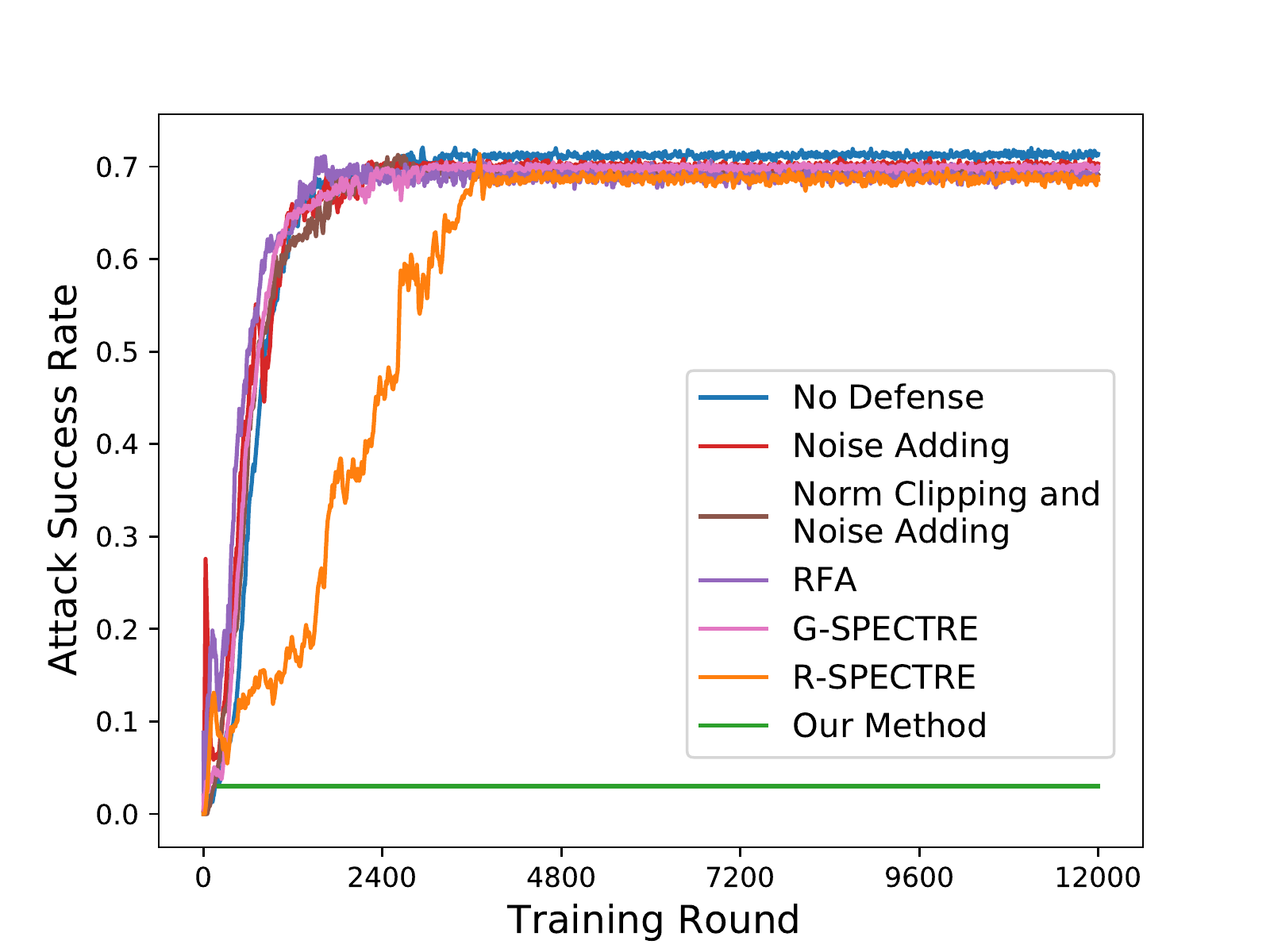}
}
\caption{Backdoor leakage causes competing defenses to fail eventually, even with a small $\alpha = 3\%$.}
\label{fig:leakage_2}
\end{figure}

Under the CIFAR-10, CIFAR-100, and Tiny-ImageNet datasets, we show that \emph{backdoor leakage} phenomenon also results in the failure of existing backdoor defenses in Figure \ref{fig:leakage_2}. Similar to Figure \ref{fig:backdoor_leakage}, we fix the malicious rate as $\alpha=3\%$ and run the experiments for $12,000$ training rounds under the static setting where data distribution is fixed over time. We can observe that the attack success rate (ASR) of all competing defenses eventually approach around $1$, $0.85$, and $0.7$ for each dataset, while the ASR of our algorithm keeps near $0$ $(0.009, 0.027, \text{and } 0.030 \text{ for each dataset})$ all the time.

\subsection{Defense under homogeneous and static clients with more datasets}
\label{app:homo_more}
We evaluate our shadow learning framework under CIFAR-100 and Tiny-ImageNet datasets in Table \ref{table:homo_CIFAR-100} and \ref{table:homo_tiny_imagenet}. We can observe that all existing defenses suffer from backdoor leakage.
CIFAR-100 and Tiny-ImageNet are also more difficult to learn compared with EMNIST dataset, and therefore the ASRs of our shadow learning framework in Table \ref{table:homo_CIFAR-100} and \ref{table:homo_tiny_imagenet} are slightly larger than those in Table \ref{table:homo_EMNIST}.

\begin{table}[htbp]
\caption{ASR for CIFAR-100 under continuous training shows the advantage of Algorithm~\ref{alg:framework}.}
\label{table:homo_CIFAR-100}
\centering
\vspace{4mm}
\begin{tabular}{ccccc}
\toprule
Defense $\setminus$
$\alpha$ & 0.15 & 0.25 & 0.35 & 0.45
\\\midrule
Noise Adding & 0.8688 & 0.8634 & 0.8525 & 0.8743
\vspace{0.8mm}\\
\vspace{0.8mm}
\makecell[c]{Clipping and\\ Noise Adding} & 0.8593 & 0.8604 & 0.8629 & 0.8688\\
RFA & 0.8642 & 0.8697 & 0.8739 & 0.8697\\
Multi-Krum & 0.8576 & 0.8594 & 0.8635 & 0.8741\\
FoolsGold & 0.8107 & 0.8251 & 0.8299 & 0.8415\\
FLAME & 0.8361 & 0.8592 & 0.8688 & 0.8691\\
CRFL & 0.8142 & 0.8197 & 0.8033 & 0.8251\\
G-SPECTRE & 0.8415 & 0.8597 & 0.8542 & 0.8673\\
R-SPECTRE & 0.7978 & 0.8033 & 0.8306 & 0.8467
\vspace{0.8mm}\\
\makecell[c]{Shadow Learning
} & \makecell[c]{\textbf{0.0765}} & \makecell[c]{\textbf{0.0929}} & \makecell[c]{\textbf{0.1694}} & \makecell[c]{\textbf{0.2247}}\\
\bottomrule
\end{tabular}
\end{table}

\begin{table}[htbp]
\caption{ASR for Tiny-ImageNet under continuous training shows the advantage of Algorithm~\ref{alg:framework}.}
\label{table:homo_tiny_imagenet}
\centering
\vspace{4mm}
\begin{tabular}{ccccc}
\toprule
Defense $\setminus$
$\alpha$ & 0.15 & 0.25 & 0.35 & 0.45
\\\midrule
Noise Adding & 0.7079 & 0.6920 & 0.7253 & 0.9139
\vspace{0.8mm}\\
\vspace{0.8mm}
\makecell[c]{Clipping and\\ Noise Adding} & 0.6839 & 0.6907 & 0.6971 & 0.7182\\
RFA & 0.7164 & 0.7206 & 0.7091 & 0.6981\\
Multi-Krum & 0.6841 & 0.7193 & 0.7032 & 0.7167\\
FoolsGold & 0.6693 & 0.6872 & 0.7013 & 0.6944\\
FLAME & 0.6767 & 0.6931 & 0.6792 & 0.6784\\
CRFL & 0.6360 & 0.6519 & 0.6440 & 0.6279\\
G-SPECTRE & 0.6423 & 0.6691 & 0.6945 & 0.6932\\
R-SPECTRE & 0.5519 & 0.6090 & 0.6495 & 0.6826
\vspace{0.8mm}\\
\makecell[c]{Shadow Learning
} & \makecell[c]{\textbf{0.0720}} & \makecell[c]{\textbf{0.0799}} & \makecell[c]{\textbf{0.1127}} & \makecell[c]{\textbf{0.1519}}\\
\bottomrule
\end{tabular}
\end{table}

\subsection{Ablation study: Main Task Accuracy (MTA) vs. Attack Success Rate (ASR) Tradeoff}
\label{app:tradeoff}

\begin{figure}[htbp]
\centering
\subfigure[$\alpha=0.45$]{
\includegraphics[width=0.46\linewidth]{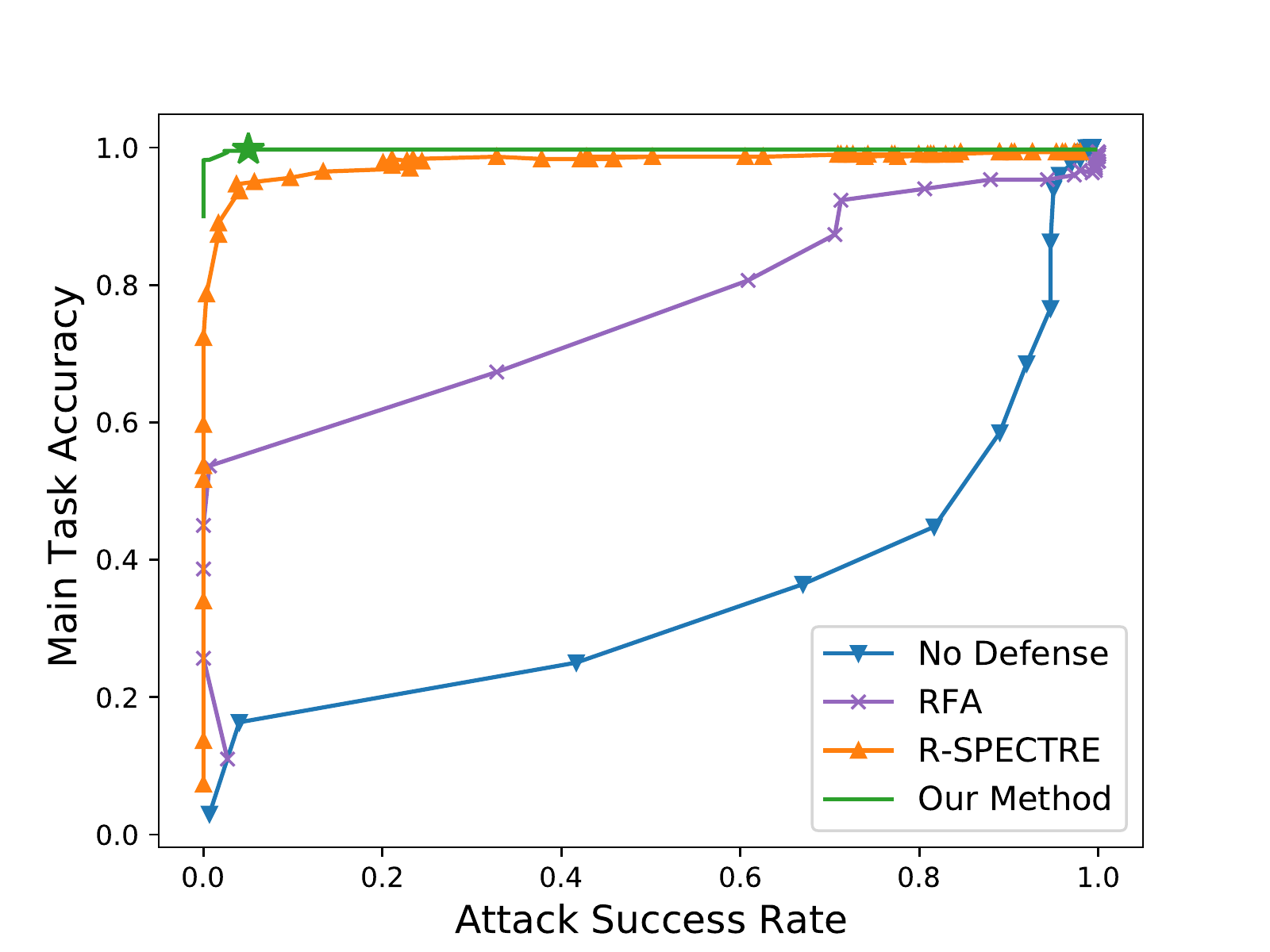}
}
\subfigure[$\alpha=0.15$]{
\includegraphics[width=0.46\linewidth]{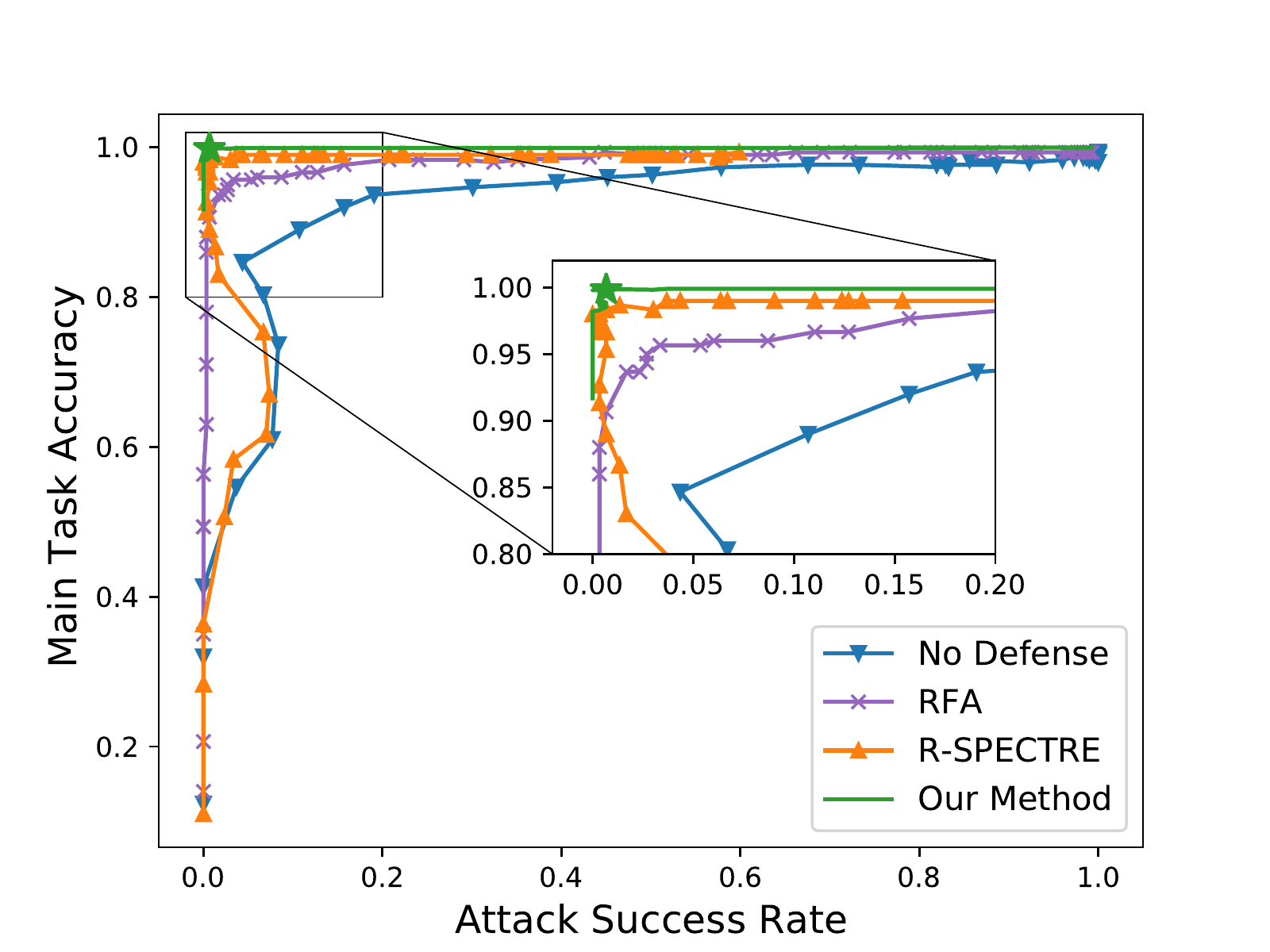}
\label{fig:tradeoff_15}
}
\caption{MTA-ASR tradeoff shows simple early stopping with no (other) defense far from the ideal (0,1). 
R-SPECTRE with early stopping is more resilient, but all early-stopping-based prior defenses suffer when clients' data changes dynamically as we show in \S\ref{section:time_varying}. 
Algorithm~\ref{alg:framework} achieves close-to-ideal tradeoffs.}
\label{fig:tradeoff}
\end{figure}


Under homogeneous and static EMNIST dataset, we run an ablation study on one of the main components: the backbone network. 
Without the backbone network, our framework (in the one-shot setting) reduces to  training baseline defenses with early stopping.
Figure~\ref{fig:tradeoff} shows the resulting achievable (ASR, MTA)  
as we tune the early-stopping round for $\alpha=0.45$ and $0.15$. 
The curves start at the bottom left (0,0) and most of them first move up, learning the main tasks. Then, the curves veer to the right as the backdoor is learned. We want algorithms that achieve points close to the top left $(0,1)$. 

When $\alpha=0.45$, Algorithm~\ref{alg:framework} achieves  
the green star. 
The blue curve (early stopping with no other defense) is far from  $(0,1)$.
This suggests that the backbone network and SPECTRE filtering are necessary to achieve the performance of Algorithm~\ref{alg:framework}. 
The curve for early-stopped RFA (purple x's) is also far from  $(0,1)$ for large $\alpha$. 
Early-stopped R-SPECTRE without the backbone network (orange triangles) achieves a good MTA-ASR tradeoff (though still worse than that of Algorithm \ref{alg:framework}). 
However, we show in \S\ref{section:time_varying} that the MTA-ASR tradeoff of R-SPECTRE is significantly worse  when clients' data distribution changes dynamically. 


When $\alpha = 0.15$, the learning rate of backdoor samples is much smaller than the main task learning rate for all curves. However, the curves for early-stopping with no defense and early-stopped RFA still cannot achieve close-to-ideal tradeoffs.

\subsection{Synthetic heterogeneous clients} 
\label{app:hetero}
In  synthetic heterogeneous EMNIST, each client receives shuffled images from $4$ randomly selected classes with $25$ samples per class. As shown in Table \ref{table:hetero_EMNIST}, it has a similar trend as the naturally heterogeneous dataset from the original EMNIST shown in Table~\ref{table:EMNIST}.

\begin{table*}[htbp]
\caption{ASR under synthetic heterogeneous EMNIST }
\label{table:hetero_EMNIST}
\centering
\vspace{4mm}
\begin{tabular}{c|ccc}
\toprule
\(\alpha\) & \makecell[c]{Our Method with\\ Sample-level Defense} & \makecell[c]{Our Method} & \makecell[c]{Our Method with\\ User-level Defense} \\\midrule
0.15 & 0.0134 & 0.0334 & 0.6756 \\
0.25 & 0.0268 & 0.0401 & 0.7424 \\
0.35 & 0.0535 & 0.0970 & 0.8127 \\
0.45 & 0.0669 & 0.1305 & 0.9833 \\
\bottomrule
\end{tabular}
\end{table*}

Further, we analyze the situation where the user-level version of our algorithm works. We construct variants of EMNIST dataset with different heterogeneity level $h$. We call the dataset $h$-heterogeneous if the first $h$-fraction of the overall training images are shuffled and evenly partitioned to each client, and for the remaining $(1-h)$-fraction samples, each client receives shuffled images from 4 randomly selected classes with $\lfloor25(1-h)\rfloor$ samples per class. For the adversarial clients, they also own 10 backdoor images. We fix the malicious rate as $\alpha=0.15\%$.
Table \ref{table:hetero_level} shows the ASR of the user-level version of our algorithm under the dataset with different heterogeneity levels.

The ASR of the user-level version of our 
algorithm is smaller than $0.1$ when $h\leq 0.4$, indicating that our user-level method can achieve the defense goal when under the dataset with low heterogeneity level.
However, when the heterogeneity level is high, the malicious clients cannot be distinguished by the aggregated statistic over all local samples, which results in the failure of the user-level defense method.

\begin{table*}[htbp]
\caption{ASR of Our Method in User-level Version with $\alpha = 0.15$}
\label{table:hetero_level}
\centering
\vspace{4mm}
\begin{tabular}{c|cccccc}
\toprule
Heterogeneity Level $h$ & 0 & 0.2 & 0.4 & 0.6 & 0.8 & 1 \\\midrule
ASR & 0.0216 & 0.0334 & 0.0969 & 0.2508 & 0.4816 & 0.6756 \\\bottomrule
\end{tabular}
\end{table*}

\subsection{More experiments under the dynamic setting} 
\label{app:dynamic} 

\begin{figure}[htbp]
\centering
\includegraphics[width=0.5\linewidth]{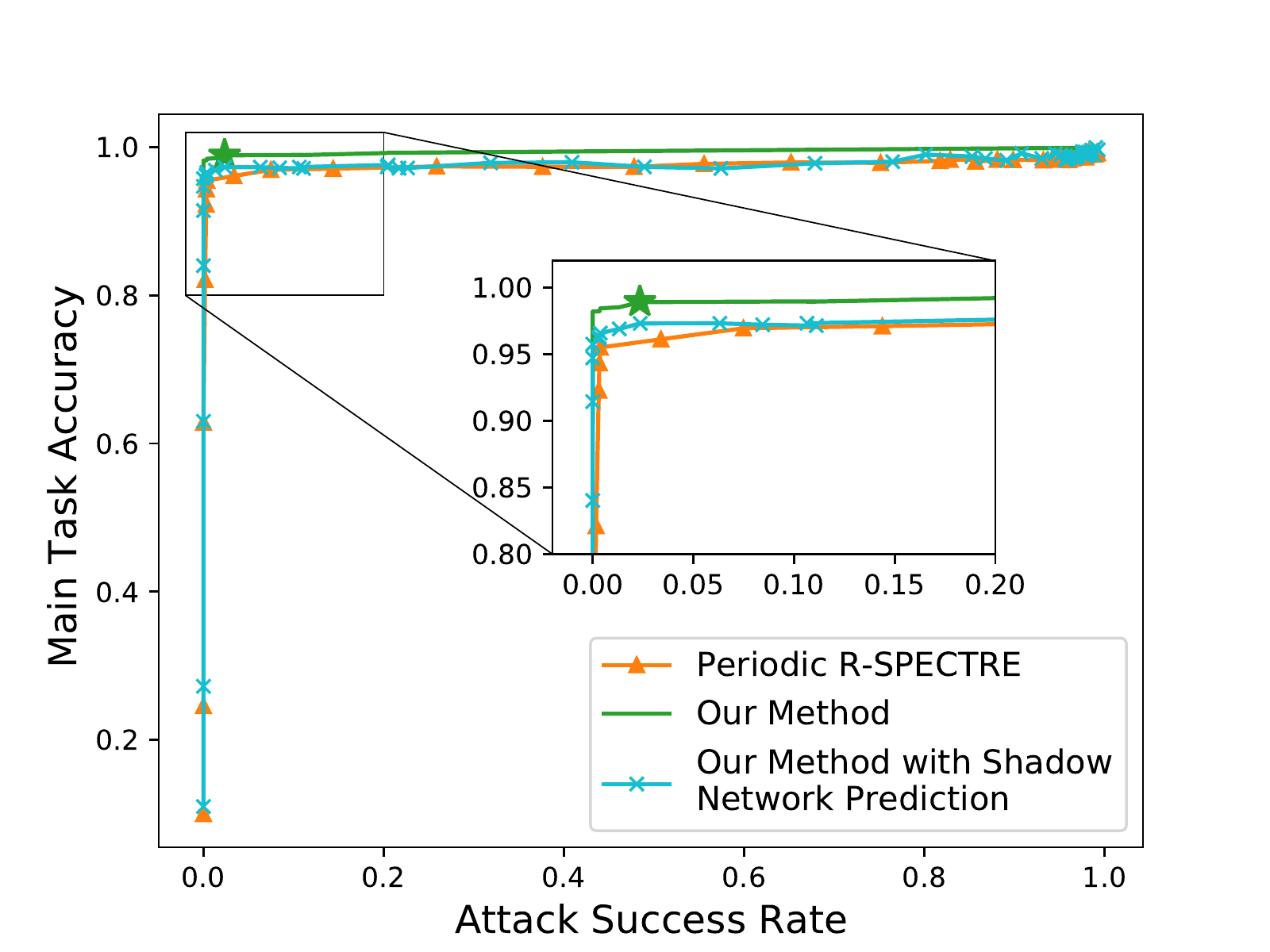}
\caption{Analysis of MTA-ASR Tradeoff with $\alpha = 0.15$ after distribution shift in phase $e3$.}
\label{fig:time_varying_e3}
\end{figure}
Recall the initial phases $e_1$ and $e_2$ in Section \ref{section:time_varying}.  In the third phase $e_3$, for all clients, we reduce the number of samples with labels $0$ and $1$ to $10\%$ of the original while keeping other samples at the same level in $e_1$, i.e., each client has one image for label $0$ and $1$ respectively, and ten images for any other label. 
This tradeoff is shown in Figure \ref{fig:time_varying_e3}.
Since the number of samples with the target label $\targetlabel = 1$ is reduced, the learning rates of both the backdoor samples and the benign samples with the target label decrease. This is similar to the setting in Figure \ref{fig:tradeoff_15} except that the proportion of samples with the target label is smaller.

\begin{figure}[htbp]
\centering
\subfigure[Phase $e_4$]{
\includegraphics[width=0.46\linewidth]{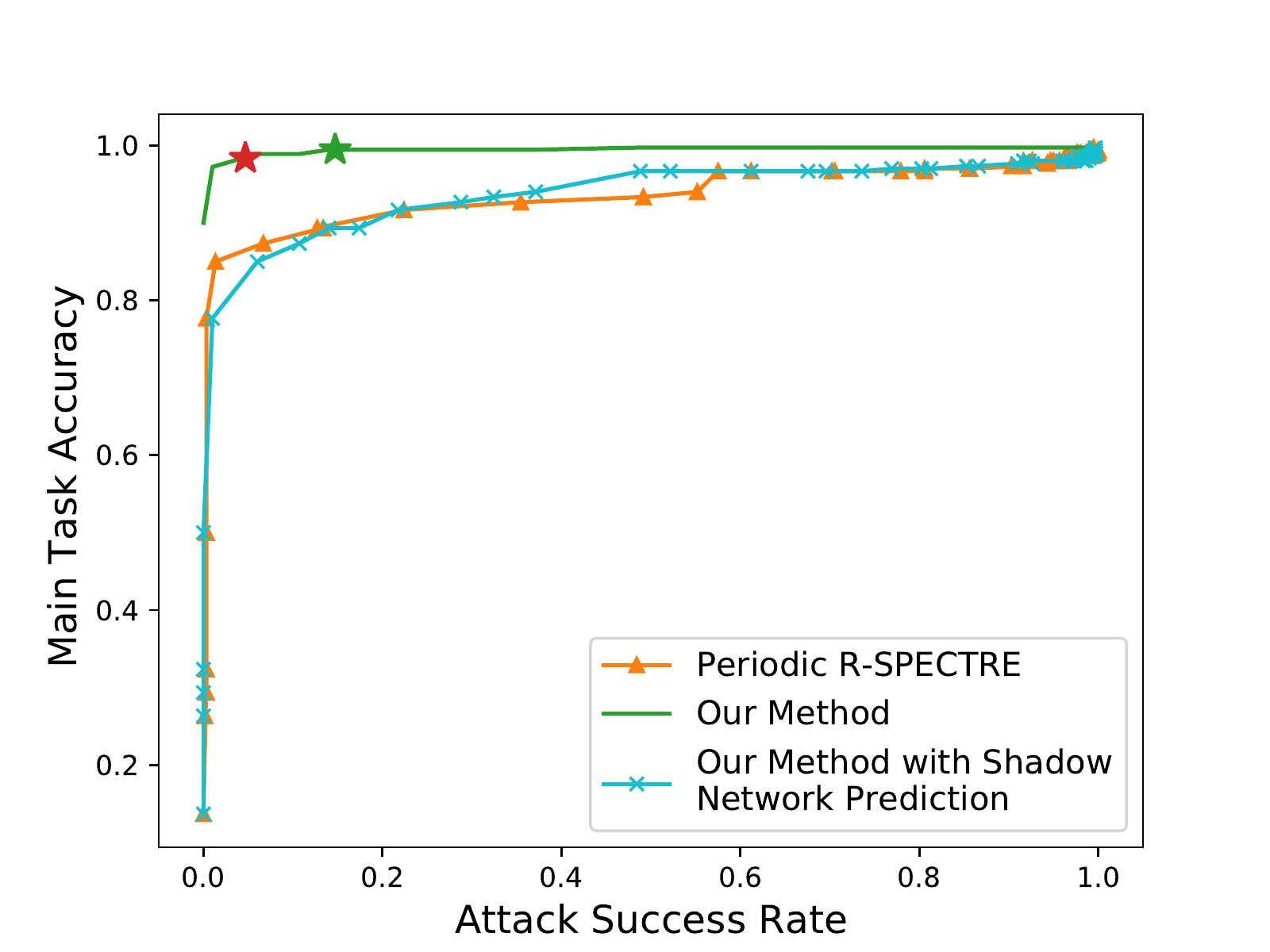}
}
\subfigure[Phase $e_5$]{
\includegraphics[width=0.46\linewidth]{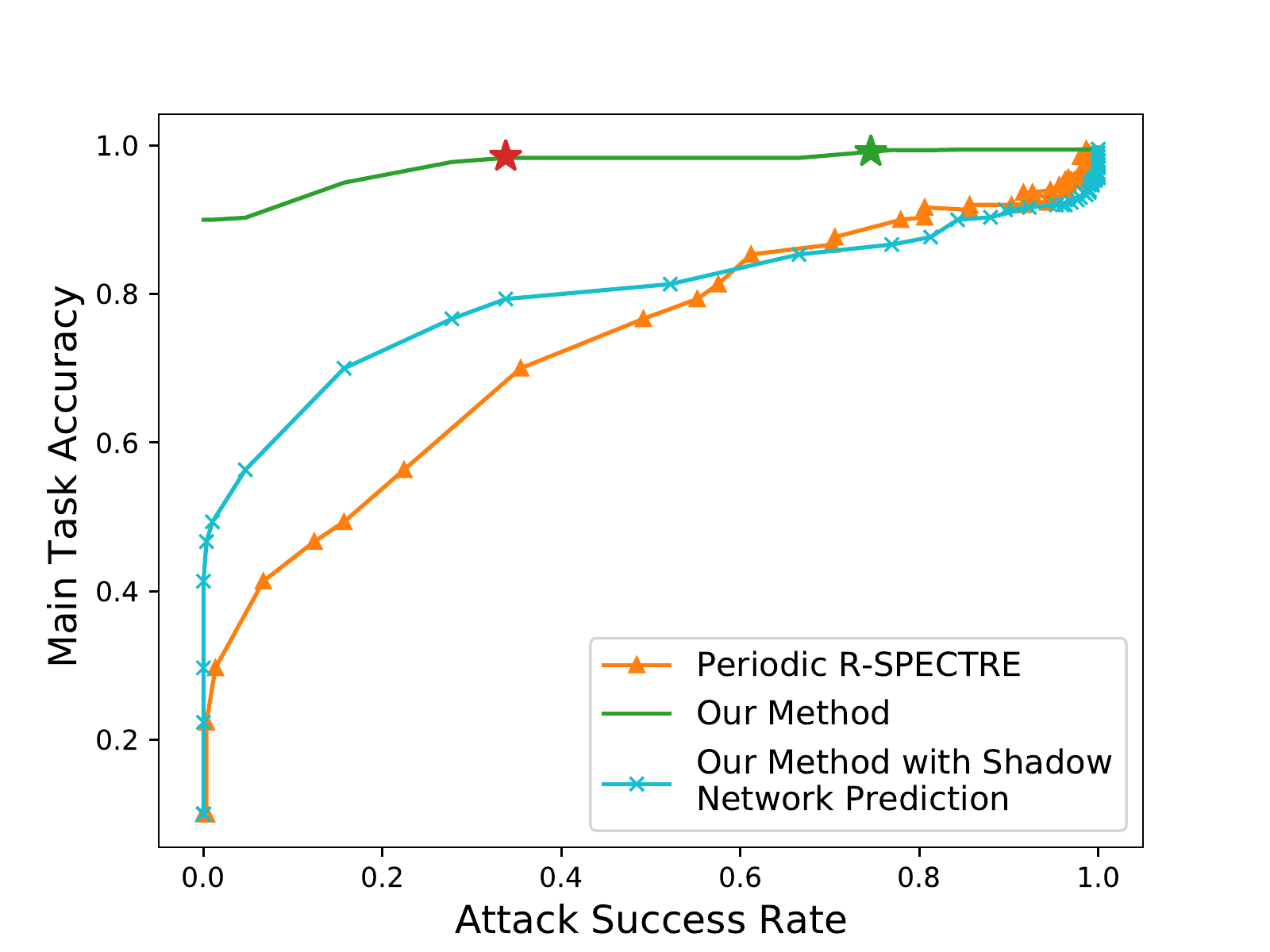}
}
\caption{Analysis of MTA-ASR Tradeoff with $\alpha = 0.15$ under Time-varying Dataset and Static Adversarial Clients}
\label{fig:time_varying_2}
\end{figure}

We also consider the case where the data distribution over malicious clients is fixed over time, which rarely happens in practice. 
In a new phase $e_4$, we let each benign client contain $5$ images for label 0 and 1 respectively, and $10$ images for any other label. 
In phase $e_5$, we further reduce the number of images each benign client contains for label $0$ or $1$ to one.
However, the local dataset of each adversarial client is always fixed as the malicious dataset in $e_1$, i.e., $10$ backdoor images in target label $\targetlabel = 1$ and $90$ images for other labels. The MTA-ASR tradeoffs of phases $e_4$ and $e_5$ are shown in Fig.\ref{fig:time_varying_2}

Our algorithm detects the data distribution change and renews the retraining period at the beginning of both phases. The MTA-ASR point our algorithm achieves is $(0.9944, 0.1471)$ and $(0.9937, 0.7458)$ in phases $e_4$ and $e_5$ respectively. The MTA-ASR tradeoffs of the comparing algorithms are much worse than ours in both phases. 

Since the benign features with label $1$ becomes more difficult to learn due to the leak of samples, and the proportion of backdoor samples among samples with target labels increases to $26\%$ and $64\%$ in phases $e_4$ and $e_5$ respectively, it is difficult to achieve the close-to-ideal MTA-ASR tradeoff.
If we increase the convergence threshold $\epsilon_2$ from $0.05\%$ to $0.5\%$ when determining the early-stopping point, we can obtain the new MTA-ASR point indicated by the red star as $(0.9834, 0.0418)$ and $(0.9834, 0.3378)$ in $e_4$ and $e_5$ respectively. By scarifying a little bit MTA (around $0.01$ in both cases), we can significantly reduce the ASR (around $0.1$ in $e_4$ and $0.4$ in $e_5$).

\subsection{Defense against adaptive malicious rate attack}
\label{app:adaptive_alpha}

We focus on an adaptive attacker strategy where the attack knows the time windows when the shadow learning framework conducts outlier detection and changes the number of malicious participating clients in each round accordingly. We assume that for every $r_0$ rounds, the attackers have the backdoor budget $B = \alpha n_{\mathcal{C}} r_0$, i.e., the adversary cannot corrupt more than $\alpha n_{\mathcal{C}} r_0$ participating clients every $r_0$ rounds, where $n_{\mathcal{C}}$ denotes the number of participating clients in each round.

\begin{figure}[htbp]
\centering
\includegraphics[width=0.5\linewidth]{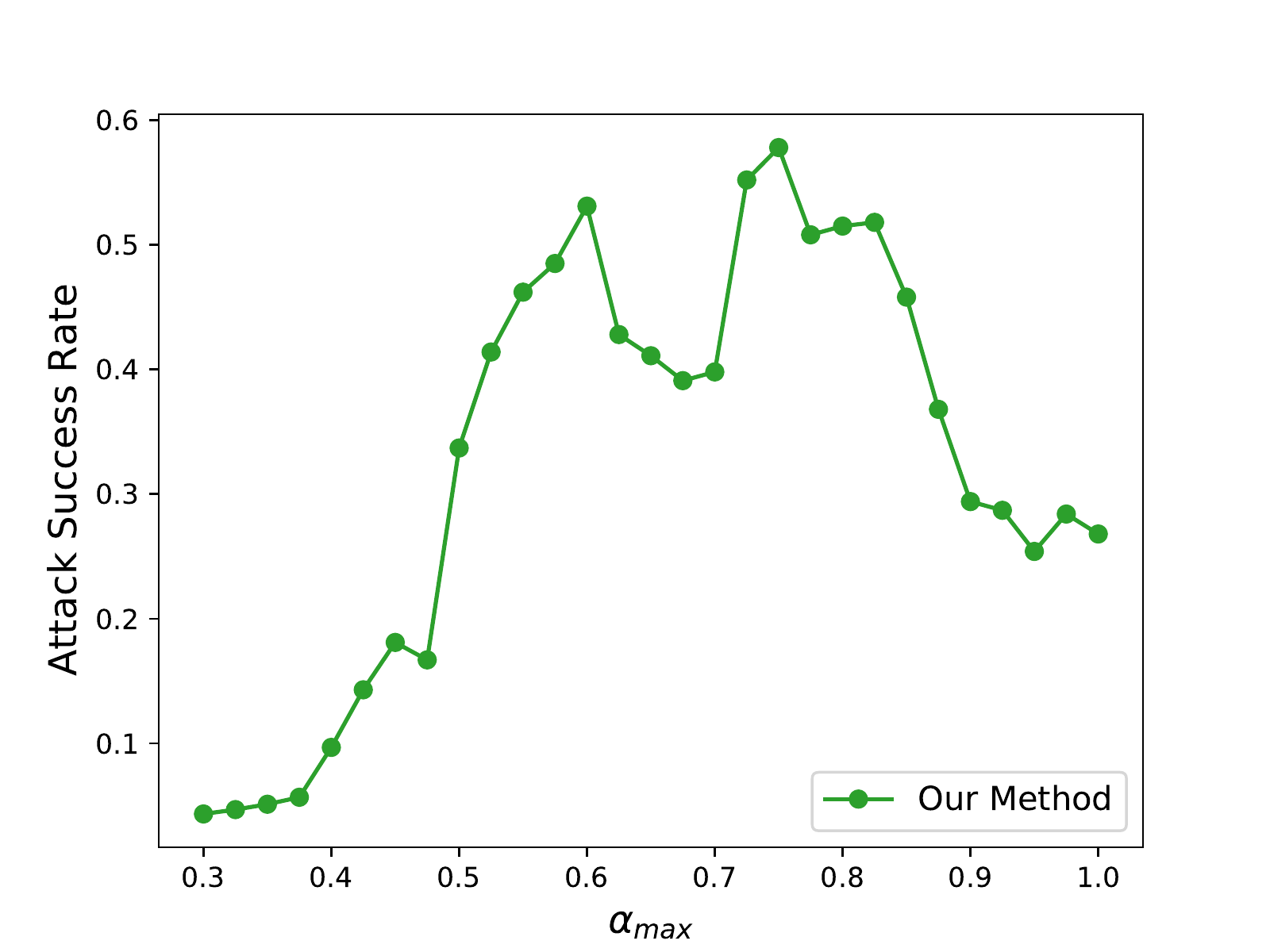}
\caption{ASR of adaptive malicious rate attack with $\alpha = 0.3$}
\label{fig:adaptive_alpha}
\end{figure}

In our experiments, we set $\alpha = 0.3$, $n_{\mathcal{C}} = 50$, $r_0 = 150$, and suppose the attacker has the knowledge that the outlier detection is conducted in a $30$-round time window.
In the time window, the attack corrupts $\alpha_{max}$-fraction participating clients, where $\alpha_{max}\geq {\alpha}$, intending to corrupt the learned SPECTRE filter. In the future $120$ rounds, the malicious rate will drop to $\alpha_{min} = \frac{5}{4}\alpha - \frac{1}{4}\alpha_{max}$ due to unlimited backdoor budget. We vary $\alpha_{max}$ from $0.3$ to $1$ and show the ASR under shadow learning framework in Figure \ref{fig:adaptive_alpha}.

As observed in Figure \ref{fig:adaptive_alpha}, the ASR is smaller than $0.05$ when $\alpha_{max}<0.4$, indicating the robustness of shadow learning framework to the adaptive malicious rate attack with moderate $\alpha_{max}$.
With the increase of $\alpha_{max}$, the SPECTRE filter will be corrupted and perform worse, and therefore, the ASR will increase when $\alpha_{max}<0.6$. However, when $\alpha_{max}$ is large enough, $\alpha_{min}$ drops significantly such that there are no sufficient malicious clients to further corrupt the model even with more corrupted filter. Therefore, the ASR then drops as the $\alpha_{max}$ grows.

\subsection{Defense against attacks with explicit evasion of anomaly detection \cite{bagdasaryan2020backdoor}}
\label{sec:fl_attack}

For attack in \cite{bagdasaryan2020backdoor}, to explicitly evade the anomaly detection, adversarial clients modify their objective (loss) function by adding an anomaly detection term $\mathcal{L}_{ano}$:
$$
\mathcal{L}_{model} = \gamma \mathcal{L}_{class} + (1-\gamma) \mathcal{L}_{ano},
$$
where $\mathcal{L}_{class}$ represents the accuracy on both the main and backdoor tasks. $\mathcal{L}_{ano}$ captures the type of anomaly detection they want to evade. In our setting, $\mathcal{L}_{ano}$ accounts for the difference between the representations of backdoor samples and benign samples with target label $\targetlabel$.
The ASR of our algorithm under such attack is shown in Table \ref{table:fl_attack} with different values of $\gamma$.

\begin{table*}[htbp]
\caption{ASR of Our Method under the Attack in \cite{bagdasaryan2020backdoor}}
\label{table:fl_attack}
\centering
\vspace{4mm}
\begin{tabular}{c|cccc}
\toprule
\(\alpha\) & $\gamma = 0.4$ & $\gamma = 0.6$ & $\gamma = 0.8$ & $\gamma = 1$ \\\midrule
0.15 & 0.0100 & 0.0100 & 0.0067 & 0.0067 \\
0.25 & 0.0133 & 0.0167 & 0.0133 & 0.0101 \\
0.35 & 0.0201 & 0.0304 & 0.0368 & 0.0312 \\
0.45 & 0.0367 & 0.0702 & 0.0635 & 0.0502 \\
\bottomrule
\end{tabular}
\end{table*}

We can observe that with different $\gamma$, the ASR of our method is always smaller than or equal to $0.07$ for the homogeneous EMNIST dataset, indicating that the attacks cannot succeed under our defense method even with explicit evasion of anomaly detection.

\subsection{Different backdoor trigger patterns}
\label{sec:trigger}

We focus on three different backdoor trigger patterns: diagonal trigger, random trigger, and periodic signal. The first two trigger patterns can be regarded as the pixel attack, while the last pattern belongs to the periodic attack \cite{barni2019new}.
The diagonal trigger consists of black pixels in the top left to bottom right diagonal, and as for the random trigger, $25$ pixels are randomly selected from the image and fixed as the trigger pattern.
For the periodic signal, we choose the sine signal with amplitude 8 and frequency of 10.
As shown in Table \ref{table:trigger}, under the homogeneous EMNIST dataset, the ASR of our method is smaller than $0.06$ in all cases, indicating our method can generalize to different types of backdoor attacks.

\begin{table*}[htbp]
\caption{ASR of Our Method under Different Trigger Patterns}
\label{table:trigger}
\centering
\vspace{4mm}
\begin{tabular}{c|ccc}
\toprule
\(\alpha\) & Diagonal Trigger & Random Trigger & Periodic Signal \\\midrule
0.15 & 0.0133 & 0.00 $\pm$ 0 & 0.00  \\
0.25 & 0.0234 & 0.0067 $\pm$ 0.001 & 0.0201 \\
0.35 & 0.0281 & 0.0268 $\pm$ 0.003 & 0.0367 \\
0.45 & 0.0569 & 0.0533 $\pm$ 0.003 & 0.0585 \\
\bottomrule
\end{tabular}
\end{table*}

\subsection{Aggregation with noise}
\label{sec:noise}

Our approach does not allow for secure aggregation, which may introduce privacy concerns. 
To this end, we consider the setting where all gradients and representations are corrupted with i.i.d. Gaussian noise $\mathcal{N}(0, \sigma^2)$, which is added to each coordinate before being uploaded from each client to the server. 
This is similar to  Differentially-Private Stochastic Gradient Descent (DP-SGD). We vary the variance $\sigma^2$ from $0.001$ to $1$, and analyze the performance of our method in terms of (MTA, ASR) pairs under the homogeneous EMNIST dataset in Table \ref{table:noise}.

\begin{table*}[htbp]
\caption{(MTA, ASR) Pair of Our Method under Different Noise Level}
\small
\label{table:noise}
\centering
\vspace{4mm}
\begin{tabular}{c|ccccc}
\toprule
\(\alpha\) & $\sigma^2 = 0$ & $\sigma^2 = 0.001$ & $\sigma^2 = 0.01$ & $\sigma^2 = 0.1$ & $\sigma^2 = 1$ \\\midrule
0.15 & $(0.997, \ 0.007)$ & $(0.998, \ 0.003)$ & $(0.994, \ 0.003)$ & $(0.983, \ 0.000)$ & $(0.873, 0.000)$  \\
0.25 & $(0.998, \ 0.010)$ & $(0.996, \ 0.013)$ & $(0.995, \ 0.007)$ & $(0.983, \ 0.000)$ & $(0.863, 0.000)$ \\
0.35 & $(0.998, \ 0.031)$ & $(0.998, \ 0.031)$ & $(0.994, \ 0.027)$ & $(0.980, \ 0.000)$ & $(0.852, 0.000)$ \\
0.45 & $(0.997, \ 0.050)$ & $(0.996, \ 0.050)$ & $(0.993, \ 0.042)$ & $(0.980, \ 0.017)$ & $(0.867, 0.000)$ \\
\bottomrule
\end{tabular}
\end{table*}

We can observe that as the noise level increases, both the MTA and ASR decrease. The intuition is that with large noise, the learning rate of the backdoor samples decreases, which makes the early-stopping framework more effective. 
However, the accuracy on main tasks also drops due to the noise. When $\sigma^2 = 0.1$, the ASR is $0$ with $\alpha \leq 0.35$, while the MTA drops from around $0.997$ to $0.98$. With $\sigma^2 = 0.01$, the added noise can reduce the ASR while keeping MTA around $0.995$.

\end{document}